\documentclass{article}

\usepackage[final]{neurips_2025}

\usepackage[utf8]{inputenc}
\usepackage[T1]{fontenc}
\usepackage{xcolor}
\definecolor{linkblue}{HTML}{0466AC}
\usepackage[colorlinks=true,linkcolor=linkblue,citecolor=linkblue,urlcolor=linkblue]{hyperref}
\usepackage{url}
\usepackage{booktabs}
\usepackage{amsfonts}
\usepackage{nicefrac}
\usepackage{microtype}

\usepackage{graphicx}
\usepackage{tabularx}
\usepackage{multirow}
\usepackage{pifont}
\usepackage{tikz}
\usetikzlibrary{positioning, arrows.meta, shapes.geometric, fit, calc, backgrounds, decorations.pathreplacing}
\usepackage{pgfplots}
\pgfplotsset{compat=1.17}
\usepackage{listings}
\usepackage{capt-of}        
\usepackage{threeparttable} 
\usepackage{longtable}

\definecolor{codegray}{rgb}{0.95,0.95,0.95}
\definecolor{paperteal}{HTML}{2B8C8C}
\definecolor{papercoral}{HTML}{D4652A}

\usepackage{xspace}
\providecommand{\eg}{e.g.\@\xspace}

\providecommand{\etc}{etc.\@\xspace}

\usepackage[capitalise,noabbrev]{cleveref}

\begin{document}

\title{Gym-V: A Unified Vision Environment System for Agentic Vision Research}

\author{
  Fanqing Meng$^{*1}$ \quad
  Lingxiao Du$^{*1}$ \quad
  Jiawei Gu$^{*1}$ \quad
  Jiaqi Liao \\
  Linjie Li$^{3}$ \quad
  Zijian Wu$^{1}$ \quad
  Xiangyan Liu$^{1}$ \quad
  Ziqi Zhao$^{4}$ \quad
  Mengkang Hu$^{5}$ \\
  Zichen Liu$^{1}$ \quad
  Jiaheng Zhang$^{1}$ \quad
  Michael Qizhe Shieh$^{1}$ \\[6pt]
  \textbf{\href{https://github.com/ModalMinds/gym-v}{https://github.com/ModalMinds/gym-v}}
}

\makeatletter
\def\@noticestring{}
\makeatother
\maketitle

\makeatletter
\renewcommand{\thefootnote}{}
\renewcommand{\@makefntext}[1]{\noindent #1}
\footnotetext{$^*$Equal contribution. \\
  $^{1}$National University of Singapore.
  $^{2}$Hong Kong University of Science and Technology.
  $^{3}$University of Washington.
  $^{4}$The Hong Kong Polytechnic University.
  $^{5}$The University of Hong Kong.
  \quad Correspond to: mengfanqing33@gmail.com}
\renewcommand{\thefootnote}{\arabic{footnote}}
\makeatother

\begin{abstract}
As agentic systems increasingly rely on reinforcement learning from verifiable rewards, standardized ``gym'' infrastructure has become essential for rapid iteration, reproducibility, and fair comparison. Vision agents lack such infrastructure, limiting systematic study of what drives their learning and where current models fall short. We introduce \textbf{Gym-V}, a unified platform of 179 procedurally generated visual environments across 10 domains with controllable difficulty, enabling controlled experiments that were previously infeasible across fragmented toolkits. Using it, we find that observation scaffolding is more decisive for training success than the choice of RL algorithm, with captions and game rules determining whether learning succeeds at all. Cross-domain transfer experiments further show that training on diverse task categories generalizes broadly while narrow training can cause negative transfer, with multi-turn interaction amplifying all of these effects. Gym-V is released as a convenient foundation for training environments and evaluation toolkits, aiming to accelerate future research on agentic VLMs.
\end{abstract}



\section{Introduction}
\label{sec:intro}

Interactive gym-style training is becoming the default for agentic LLMs. Platforms such as GEM~\citep{GEM} and Reasoning Gym~\citep{reasoninggym} have matured into large procedural suites with unified workflows. Their success rests on three recurring ingredients: \textit{procedural generation}, \textit{automatic verification}, and \textit{standardized interfaces}. This combination scales training and supports meaningful, apples-to-apples comparisons. Vision agents do not yet benefit from a similarly unified platform for scalable training and fair comparison.

\begin{table}[t]
  \caption{Comparison of gym-style environment platforms.
    $\bullet$\,=\,supported, $\circ$\,=\,not supported.
    ST: single-turn; MT: multi-turn; MA: multi-agent; Eval: evaluation wrapper; Off.: offline data; Tool: tool-augmented; Diff.: difficulty levels; MM: multimodal; Eaas: evaluation-as-a-service.}
  \label{tab:comparison}
  \centering
  \setlength{\tabcolsep}{3.5pt}
  \footnotesize
  \begin{tabular}{@{}lc cccccc ccc@{}}
    \toprule
    & & \multicolumn{6}{c}{\textbf{Interaction Workflows}} & \multicolumn{3}{c}{\textbf{Infra.}} \\
    \cmidrule(lr){3-8} \cmidrule(lr){9-11}
    Platform & \#Env & ST & MT & MA & Eval & Off. & Tool & Diff. & MM & Eaas \\
    \midrule
    \multicolumn{11}{l}{\textit{Text-centric}} \\
    TextArena~\citep{textarena}        &  74  & $\circ$ & $\bullet$ & $\bullet$ & $\circ$ & $\circ$ & $\circ$ & $\circ$ & $\circ$ & $\circ$ \\
    Reasoning Gym~\citep{reasoninggym} & 100+ & $\bullet$ & $\circ$ & $\circ$ & $\circ$ & $\circ$ & $\circ$ & $\bullet$ & $\circ$ & $\circ$ \\
    RLVE~\citep{rlve}                  & 400  & $\bullet$ & $\circ$ & $\circ$ & $\circ$ & $\circ$ & $\circ$ & $\bullet$ & $\circ$ & $\circ$ \\
    GEM~\citep{GEM}                    &  24  & $\bullet$ & $\bullet$ & $\bullet$ & $\bullet$ & $\bullet$ & $\bullet$ & $\bullet$ & $\circ$ & $\circ$ \\
    \midrule
    \multicolumn{11}{l}{\textit{Vision-centric}} \\
    VisGym~\citep{visgym}              &  17  & $\circ$ & $\bullet$ & $\circ$ & $\circ$ & $\circ$ & $\circ$ & $\bullet$ & $\bullet$ & $\circ$ \\
    VAGEN~\citep{vagen}                &   5  & $\circ$ & $\bullet$ & $\circ$ & $\circ$ & $\circ$ & $\circ$ & $\circ$ & $\bullet$ & $\circ$ \\
    VLM-Gym~\citep{vlmgym}             &   4  & $\circ$ & $\bullet$ & $\circ$ & $\circ$ & $\circ$ & $\circ$ & $\bullet$ & $\bullet$ & $\circ$ \\
    KorGYM~\citep{shi2025korgym}              &  50+ & $\bullet$ & $\bullet$ & $\circ$ & $\bullet$ & $\circ$ & $\circ$ & $\bullet$ & $\bullet$ & $\circ$ \\
    \midrule
    \textbf{Gym-V (ours)}             & \textbf{179} & $\bullet$ & $\bullet$ & $\bullet$ & $\bullet$ & $\bullet$ & $\bullet$ & $\bullet$ & $\bullet$ & $\bullet$ \\
    \bottomrule
  \end{tabular}
\end{table}

\begin{figure*}[tb]
  \centering
  \includegraphics[width=0.95\linewidth]{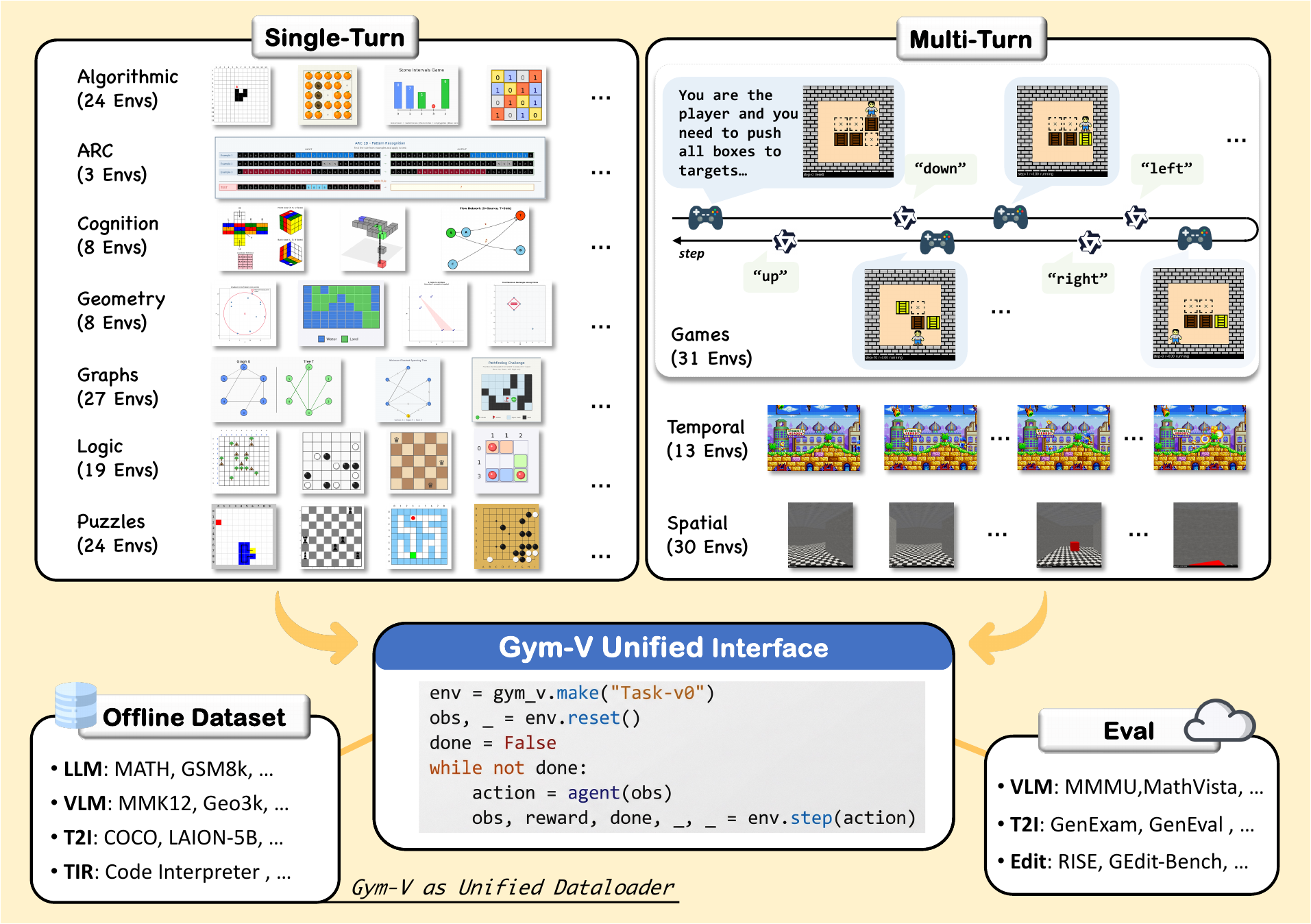}
  \caption{Overview of Gym-V. \textbf{Top:} 105 single-turn and 74 multi-turn environments across 10 categories. \textbf{Bottom:} a unified \texttt{reset}/\texttt{step} interface shared by interactive environments, offline datasets, and evaluation benchmarks.}
  \label{fig:overview}
\end{figure*}

Moving from language to pixels raises the bar: perception becomes a bottleneck that amplifies difficulty beyond text-only analogues despite similar reasoning demands~\citep{visgym}. Moreover, how visual observations are framed and contextualized for the model can determine whether reinforcement learning succeeds at all, making observation design a critical yet underexplored variable in vision-agent training. Studying this variable at scale requires diverse visual tasks under consistent interfaces, yet existing platforms such as VAGEN~\citep{vagen} and VLM-Gym~\citep{vlmgym} cover only a few task families with limited workflow support. As \Cref{tab:comparison} shows, text suites provide broad workflow support but omit pixels, while vision gyms include images but cover far fewer workflows, highlighting the need for a unified platform that spans both settings.


We introduce \textbf{Gym-V} to fill this gap with a gym-style platform that standardizes how vision agents are trained and evaluated (\cref{fig:overview}). A Gym-compatible API~\citep{1606.01540} runs the same loop on a single-turn visual puzzle or a multi-turn game episode, while built-in verifiers check correctness automatically. Gym-V spans 179 environments across 10 domains (\cref{tab:env-overview}), and the same API unifies them with offline datasets and evaluation benchmarks under one interface, including a distributed reward service for generative tasks. To make observation design an experimentally controllable variable, a composable wrapper layer controls what context the agent receives without modifying the underlying environment.


With observation design as a controllable variable, Gym-V enables a unified investigation of what drives vision-agent performance. Zero-shot evaluation of nine VLMs reveals large capability gaps across model generations, as a newer 32B model outperforms a conventional 72B model by 1.8$\times$ and even the strongest model leaves substantial headroom. To understand what drives these gaps, we train with GRPO~\citep{guo2025deepseek,shao2024deepseekmath}, GSPO~\citep{zheng2025gspo}, and SAPO~\citep{gao2025sapo} and find that \textbf{observation scaffolding is the most decisive factor for learning success}. Adding captions consistently improves learning across all tested environments and removing game rules can prevent learning altogether, while \textbf{no single RL algorithm dominates across all settings}: the best choice is scenario-dependent, with differences most visible in long-horizon multi-turn stability. This sensitivity to task presentation extends to curriculum design, where a cross-domain transfer study shows that \textbf{training on categories with diverse sub-skills transfers broadly} while narrow training can cause negative transfer. Multi-turn interaction amplifies all of these effects, making both scaffolding and curriculum choices increasingly critical over longer horizons. Our contributions are:

\begin{enumerate}
  \item \textbf{A unified gym-style platform for vision agents.}
    Gym-V provides a Gym-compatible API that unifies interactive training, offline supervision, and benchmark evaluation under one interface. Composable wrappers make observation design an explicit experimental variable, and a distributed reward service extends scoring to generative tasks.
  \item \textbf{A diverse suite of vision environments.}
    The suite spans 10 categories with procedural generation and difficulty presets, providing the scale and control for curriculum learning and cross-domain generalization studies.
  \item \textbf{Empirical insights on vision-agent training.}
    Controlled experiments show that observation scaffolding is the most decisive factor for vision-agent training, more consequential than the choice of RL algorithm. We additionally find that diverse curricula generalize across domains while narrow training can hurt, and that multi-turn interaction amplifies all of these effects.
\end{enumerate}

\section{Related Work}
\label{sec:related}

\noindent\textbf{VLM Benchmarks and Interactive Environments.}
Static benchmarks for vision-language models---from visual question answering~\citep{marino2019ok,gurari2018vizwiz} to comprehensive suites covering mathematical reasoning~\citep{lu2023mathvista,wang2024measuring}, chart understanding~\citep{mathew2021docvqa,masry2022chartqa}, and multi-discipline problems~\citep{yue2024mmmu,liu2024mmbench}---have driven rapid progress but share fundamental limitations: fixed datasets are susceptible to contamination, evaluation is one-shot, and no training signal is provided for reinforcement learning.
Interactive benchmarks extend evaluation to web-based GUI navigation~\citep{zhou2023webarena,anupam2025browserarena,deng2023mind2web,koh2024visualwebarena} and game-based environments~\citep{textarena,hu2025lmgame}, yet each defines its own interface and covers narrow task families, preventing unified training pipelines.
Gym-V subsumes static evaluation via its single-turn mode while natively supporting multi-turn interaction under one consistent interface.

\noindent\textbf{Gym-style Frameworks for Agent Training.}
The success of reinforcement learning from verifiable rewards (RLVR)~\citep{lambert2024tulu,guo2025deepseek} has motivated gym-style frameworks that standardize procedural task generation with automatic verification.
Reasoning Gym~\citep{reasoninggym} provides 100+ text-based reasoning environments with parametric difficulty control.
GEM~\citep{GEM} adopts a transition-wise design supporting per-step rewards, tool-augmented interaction, and offline datasets.
Both operate exclusively in the text domain and do not address the additional complexity of visual observations.
For vision, VLM-Gym~\citep{vlmgym} and VAGEN~\citep{vagen} propose interactive visual environments but cover only 4 and 5 tasks respectively;
KorGYM~\citep{shi2025korgym} offers 50+ games with partial visual support but lacks unified RL training integration and tool-augmented workflows.
\Cref{tab:comparison} provides a systematic comparison.
Gym-V extends the gym paradigm to vision with 179 environments, an evaluation-as-a-service architecture, and native support for six interaction workflows---single-turn, multi-turn, multi-agent, tool-augmented, VLM evaluation, and image generation model evaluation---under one API.


\section{Gym-V System}
\label{sec:system}

Gym-V is a unified environment system for vision-centric learning and evaluation.
We describe the system in mainly three parts: the core environment interface (\cref{sec:env-interface}), the wrapper and configuration layer (\cref{sec:wrappers}), and the distributed reward service (\cref{sec:reward-service}).

\subsection{Environment Interface}
\label{sec:env-interface}

Gym-V follows a Gym-compatible API aligned with the multi-agent interface of Ray RLlib~\citep{liang2018rllib,liang2021rllib}. Each environment implements the standard \texttt{reset()} and
\texttt{step()} functions. A typical interaction loop is shown in Figure \ref{fig:overview}:


\paragraph{Multi-agent support.}
All observations, actions, rewards, and termination signals are structured as dictionaries indexed by agent identifiers, following the Ray RLlib multi-agent convention. It enables seamless support for single-agent, competitive, and cooperative settings without API changes.

\paragraph{Offline dataset support.}
Gym-V also supports offline training under the same environment abstraction. In online RL, each training iteration samples
\texttt{batch\_size} independent multi-turn games, where each game consists of multiple interaction steps. In contrast, for offline tasks such as VQA, each iteration samples a single-turn game involving \texttt{batch\_size} independent instances, each terminating after one step. Both settings are treated uniformly as batched episodes, enabling the same training pipeline to operate across online and offline settings.

\subsection{Wrappers as Experimental Variables} \label{sec:wrappers}

Gym-V provides a composable \emph{wrapper} subsystem that intercepts the agent--environment interaction boundary and transforms the presentation and evaluation of a task—without modifying the underlying environment dynamics. Wrappers can rewrite task rules/descriptions, augment observations (e.g., state summaries), control the context provided to the agent via configurable history windows, parse/validate agent outputs before execution; external tools (e.g., Python interpreters) are implemented as wrappers that convert tool-call actions into observations without consuming environment steps.
By treating rules, captions, and context as first-class wrapper interventions, Gym-V makes task representation an explicit experimental variable. We use wrapper to perform controlled ablations over rule phrasing, caption availability, and history length for long-horizon vision reasoning. Since wrappers compose and are logged in the environment specification, these interventions are reproducible and comparable across environments under identical dynamics.

\subsection{Evaluation-as-a-Service and Unified Evaluation}
\label{sec:reward-service}

Interactive environments use built-in rule-based verifiers (\eg, checking Sudoku validity or game win conditions) that run locally with negligible overhead. 
However, generative tasks---such as text-to-image synthesis or image editing---require learned reward models (CLIP~\citep{radford2021learningclip}, HPSv3~\citep{ma2025hpsv3}, \etc) that are expensive to install, configure, and run, and that benefit from batched GPU inference.

Gym-V addresses this with a \textbf{distributed reward service} deployed via Ray Serve, following the \emph{evaluation-as-a-service} (EaaS) paradigm. 
From the client's perspective, scoring a generated image requires only an HTTP request:

\begin{lstlisting}[language=Python]
client = BaseNetworkClient(
    endpoint=f"{server_url}/v1/generate"
)
response = client.request({
    "model": "geneval",
    "prompt": "a red car to the left of a blue bus",
    "multimodal_outputs": {
        "image": "<base64>"
    },
    "metadata": { ... }
})
\end{lstlisting}

The service manages model loading, automatic batching, and horizontal scaling across GPUs, and hosts heterogeneous reward backends behind a unified scoring API. 
These may include locally deployed models (e.g., CLIP~\citep{radford2021learningclip}, Dino~\citep{caron2021emergingdino}), VLMs served via vLLM~\citep{kwon2023efficient}, or external closed-source models accessed through APIs. 
Swapping reward models only requires changing the service configuration without modifying any environment or training code.

Building on this architecture, Gym-V unifies benchmark evaluation under the same interaction protocol used for RL. 
Each benchmark is wrapped as an environment exposing the same interface: discriminative VLM benchmarks are integrated through existing toolkits such as VLMEvalKit~\citep{duan2024vlmevalkit}, while generative benchmarks (text-to-image and image editing) delegate scoring to the evaluation service described above. 
This design removes separate evaluation pipelines and enables both discriminative and generative tasks to be evaluated through a single interface.

\begin{figure}[tb]
  \centering
  \includegraphics[width=\linewidth]{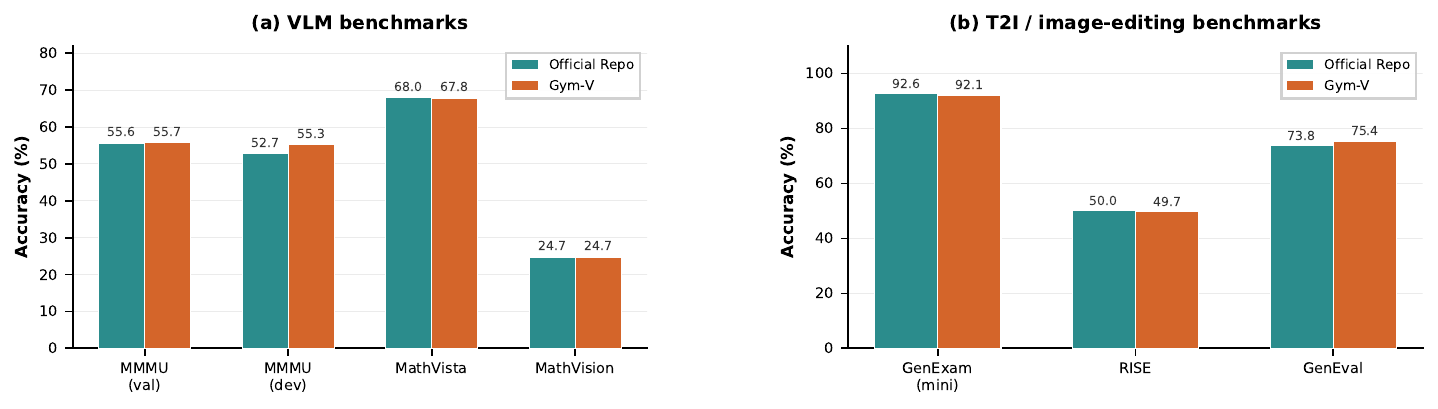}
  \caption{Evaluation fidelity of Gym-V against official pipelines.
    \textbf{(a)}~VLM evaluation on Qwen2.5-VL-7B-Instruct reproduces official VLMEvalKit scores.
    \textbf{(b)}~Text-to-image and image-editing evaluation closely matches official GenExam, RISE, and GenEval results.
  }
  \label{fig:eval-vlm}
\end{figure}

We validate the fidelity of this abstraction by comparing Gym-V with official benchmark pipelines, showing near-identical VLM scores and closely matched results on generative benchmarks including GenExam~\citep{wang2025genexam}, RISE~\citep{zhao2025envisioning}, and GenEval~\citep{ghosh2023geneval} (\cref{fig:eval-vlm}).

\subsection{Environment Suite and Taxonomy}
\label{sec:environments}

Gym-V ships an environment suite of 202 image-based interactive tasks with a shared \texttt{reset}/\texttt{step} interface and deterministic seeding. This unified interface, together with broad training-environment coverage, standardizes model IO across tasks, making cross-environment evaluation and training directly comparable and enabling systematic \textbf{transfer studies} without task-specific glue code. Per-environment details and example observations are provided in the supplementary material.

\paragraph{Design principles.}
Gym-V environments are constructed under three principles.
\textbf{Visual Grounding:} all observations are rendered as images rather than structured text; task-relevant information (spatial relations, object states, layout) is encoded visually, requiring perceptual reasoning instead of text pattern matching.
\textbf{Automatic Verification with Large Solution Spaces:} all tasks admit deterministic verification, and are designed with large sets of valid solutions so reward is not tied to a narrow answer format, reducing reward hacking and brittle overfitting.
\textbf{Parametric and Scalable Difficulty:} each environment exposes parameters controlling complexity (e.g., grid size, object count, solution depth), allowing difficulty to scale with model capability for continual evaluation.

\paragraph{Taxonomy and coverage.}
As summarized in \cref{tab:env-overview}, Gym-V spans 10 categories covering both single-turn visual reasoning and multi-turn sequential decision making: single-turn tasks evaluate perception and structured inference from static observations, while multi-turn tasks test strategy and long-horizon control under evolving states. The same abstraction also supports offline dataset training (one-step supervision or trajectory-based learning) under the identical step semantics.

\paragraph{Difficulty control and distribution shifts.}
Each environment provides a \texttt{DIFFICULTY} dictionary mapping integer levels (0, 1, 2) to predefined parameter configurations (level 0 as default), offering a standardized knob for difficulty sweeps and \textbf{cliff analyses} that reveal sharp performance breakpoints as complexity increases.

\begin{table*}[!t]
  \caption{Overview of the Gym-V environment suite (179 environments across 10 categories).
  All environments share the \texttt{reset}/\texttt{step} interface and support
  deterministic seeding.}
  \label{tab:env-overview}
  \centering

  {\footnotesize                     
  \setlength{\tabcolsep}{3pt}        
  \renewcommand{\arraystretch}{0.82} 
  \setlength{\abovecaptionskip}{2pt}
  \setlength{\belowcaptionskip}{-6pt}

  \begin{tabularx}{\textwidth}{@{}l c X@{}}
  \toprule
  Category & \#Envs & Description \\
  \midrule

  \multicolumn{3}{@{}l}{\textbf{Single-Turn (105)}} \\
  \addlinespace[1pt]

  Algorithmic & 21 & Grid-based tasks that require algorithmic reasoning---BFS
  traversal, cellular automata simulation, and game-theoretic
  computation---over visually rendered grids (e.g.\ \textit{GridBFS}). \\
  ARC & 3 & Abstract pattern induction tasks that challenge models to infer
  transformation rules from few visual input--output grid pairs and generalize
  to unseen instances (e.g.\ \textit{ArcAgi}). \\
  Cognition & 10 & Perceptual inference tasks involving 3D mental rotation,
  visual pattern completion, and spatial reasoning from rendered scenes
  (e.g.\ \textit{RubiksCube-QA}). \\
  Geometry & 8 & Geometric reasoning tasks requiring computation of convex
  hulls, enclosing circles, and areas over visually rendered point sets and
  shapes (e.g.\ \textit{ConvexHull}). \\
  Graphs & 23 & Graph-theoretic reasoning tasks including traversal,
  shortest-path computation, spanning trees, coloring, and matching over
  visually rendered graphs (e.g.\ \textit{ShortestPath}). \\
  Logic & 17 & Constraint satisfaction and logical deduction on grid-based
  puzzles, requiring systematic elimination and inference (e.g.\
  \textit{MiniSudoku}). \\
  Puzzles & 23 & Multi-step combinatorial puzzles requiring planning, state
  tracking, and search over visual representations (e.g.\
  \textit{TowerOfHanoi}). \\

  \midrule
  \multicolumn{3}{@{}l}{\textbf{Multi-Turn (74)}} \\
  \addlinespace[1pt]

  Games & 31 & Strategic decision-making in board, card, and tile games,
  covering both single-player optimization and multi-player adversarial play
  (e.g.\ \textit{Chess}). \\
  Spatial & 30 & Goal-directed navigation and object manipulation in 2D
  grid-worlds and 3D first-person environments, testing spatial reasoning and
  planning (e.g.\ \textit{DoorKey}). \\
  Temporal & 13 & Reactive control in retro arcade games with continuous
  frame-based state evolution, demanding real-time visual perception and
  action selection (e.g.\ \textit{StreetsOfRage2}). \\

  \bottomrule
  \end{tabularx}
  }
\end{table*}

\subsection{Performance Across Categories}
\label{sec:bench-performance}


\begin{figure*}[!t]
  \centering

  \captionof{table}{%
    \textbf{Left:} Single-turn evaluation ($\times 100$) across 7 domain categories .
    \textbf{Right:} Multi-turn evaluation ($\times 100$) across Games (12 envs), Spatial/2D (6 Minigrid envs), and Spatial/3D (3 MiniWorld envs).
    \textbf{Avg:} mean over all 10 columns reported in this table.
    Best result per column in \textbf{bold}.
    Note that Minigrid (Sp./2D) uses shaped episodic returns that include negative penalties (e.g., stepping into hazards), so mean@3 returns can be below zero.
  }
  \label{tab:benchmark}

  \begin{threeparttable}
    \resizebox{\linewidth}{!}{%
      \begin{tabular}{@{}l ccccccc @{\hspace{8pt}} ccc @{\hspace{8pt}} c@{}}
        \toprule
        & \multicolumn{7}{c@{\hspace{8pt}}}{Single-turn} & \multicolumn{3}{c@{\hspace{8pt}}}{Multi-turn} & Avg \\
        \cmidrule(lr){2-8} \cmidrule(lr){9-11}
        Model
        & \rotatebox{70}{Algo.}
        & \rotatebox{70}{ARC}
        & \rotatebox{70}{Cogn.}
        & \rotatebox{70}{Geom.}
        & \rotatebox{70}{Graphs}
        & \rotatebox{70}{Logic}
        & \rotatebox{70}{Puzzles}
        & \rotatebox{70}{Games}
        & \rotatebox{70}{Sp./2D}
        & \rotatebox{70}{Sp./3D}
        & \rotatebox{70}{Avg} \\
        \midrule
        Gemini-3-Pro       & \textbf{94.9} & \textbf{40.1} & \textbf{64.9} & \textbf{81.0} & \textbf{76.5} & \textbf{79.3} & \textbf{82.9} & \textbf{64.3} & \textbf{60.9} & \textbf{86.1} & \textbf{73.1} \\
        Qwen3-VL-Plus      & 75.0 & 16.0 & 48.0 & 50.9 & 51.4 & 62.2 & 57.8 & 17.8 & $-$13.6 & 63.7 & 42.9 \\
        Claude Sonnet 4.5  & 60.7 & 17.8 & 39.7 & 41.3 & 41.2 & 65.7 & 56.3 & 19.3 & $-$16.7 & 43.2 & 36.9 \\
        Qwen3-VL-32B       & 63.0 & 8.2 & 44.6 & 43.7 & 47.6 & 62.8 & 41.3 & 24.7 & $-$16.7 & 43.1 & 36.2 \\
        GPT-5.2            & 34.9 & 9.8 & 37.6 & 40.0 & 38.6 & 33.7 & 34.6 & 21.1 & $-$16.7 & 64.5 & 29.8 \\
        Qwen3-VL-8B        & 35.0 & 3.7 & 24.8 & 19.0 & 21.9 & 40.3 & 22.8 & 4.6 & $-$16.7 & 9.5 & 16.5 \\
        Qwen2.5-VL-72B     & 28.5 & 6.6 & 26.5 & 24.5 & 27.7 & 30.8 & 25.2 & 5.5 & $-$16.7 & 73.5 & 23.2 \\
        Qwen2.5-VL-32B     & 22.9 & 4.4 & 23.5 & 19.2 & 24.8 & 23.5 & 26.6 & 12.2 & $-$6.6 & 63.3 & 21.4 \\
        Qwen2.5-VL-7B      & 8.2 & 2.8 & 11.7 & 8.6 & 13.5 & 8.6 & 9.6 & 4.1 & $-$11.1 & 10.5 & 6.6 \\
        \bottomrule
      \end{tabular}%
    }
  \end{threeparttable}

  \vspace{6pt} 

  \includegraphics[width=\linewidth]{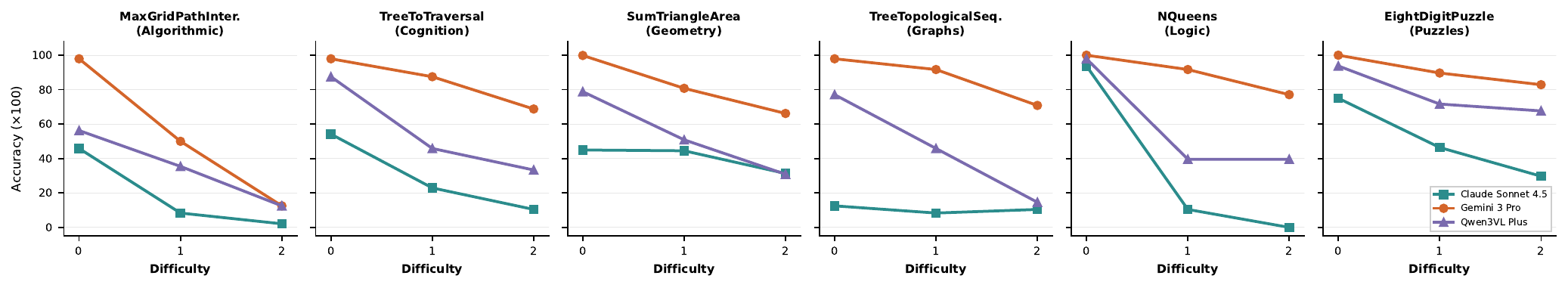}
  \captionof{figure}{Accuracy ($\times 100$) vs.\ difficulty level on six representative environments from different categories.}
  \label{fig:difficulty-cliff}
\end{figure*}

\Cref{tab:benchmark} reports zero-shot mean@3 scores on 9 models~\citep{team2023gemini,Qwen2.5-VL,Qwen3-VL,yang2025qwen3} across the 7 single-turn categories (left) and multi-turn results under a memoryless setting (context window size\,=\,0, right), where each step receives only the current observation with no conversation history; \textbf{Avg} averages all columns reported in the table. Note that we find that none of the models could complete the video task within a certain cost, and we analyze this in the supplementary materials.

\textbf{Model hierarchy.}
Performance shows a clear ordering across model classes. Gemini-3-Pro achieves the strongest overall score (Avg\,=\,73.1), while the best open-weight model (Qwen3-VL-32B) reaches 36.2, leaving a sizable gap between closed and open models. Within open-weight models, newer reasoning-oriented training appears to matter more than raw scale: averaged over the seven single-turn categories, Qwen3-VL-32B reaches 44.5, which is $\approx$1.8$\times$ Qwen2.5-VL-72B (24.3), suggesting improved training recipes can yield outsized gains for visual reasoning.

\textbf{Domain difficulty is uneven.}
Some categories are substantially harder than others. In particular, \textbf{ARC remains challenging for all models}: even Gemini-3-Pro achieves only 40.1, indicating that abstract visual pattern reasoning is far from solved. Other categories exhibit higher scores but still show wide model separation, implying that Gym-V is not saturated even in easier domains.

\textbf{Scale helps, but does not close the gap.}
Within the Qwen2.5-VL family, scaling from 7B to 72B improves the single-turn category average from 9.0 to 24.3 (2.7$\times$), but still falls well below the best closed-source result, highlighting remaining headroom for better training and reasoning supervision.

\textbf{The difficulty cliff (degradation rates).}
Beyond zero-shot performance, \Cref{fig:difficulty-cliff} reveals a consistent \emph{difficulty cliff}. We quantify degradation with a robustness ratio as $\rho=\mathrm{Acc}(d{=}2)/\mathrm{Acc}(d{=}0)$: on \texttt{MaxGridPathIntersection}, $\rho$ drops to 0.13 (Gemini-3-Pro), 0.22 (Qwen3-VL-Plus), and 0.05 (Claude), while on \texttt{NQueens} it is 0.77 (Gemini-3-Pro), 0.40 (Qwen3-VL-Plus), and 0.00 (Claude-Sonnet-4.5), indicating that some tasks collapse rapidly as complexity increases even for frontier models. Such steep cliffs highlight the largest capability gaps; unlike static offline datasets that may become saturated and less suitable for RL as models improve, Gym-V can continually raise difficulty to preserve a non-trivial learning signal, and the steepest cliffs mark the highest potential return for RL training and curriculum learning. These gaps motivate the training experiments below.       

\section{Empirical Studies With Gym-V}

\label{sec:training}

We demonstrate how Gym-V facilitates reinforcement learning research in vision-centric and agentic settings through a series of empirical studies.
Given practical training budgets, we focus on single-turn and multi-turn game environments, which allow efficient yet representative RL experimentation. 
We first present algorithm comparison results (\cref{sec:train-algo}), followed by multi-turn and vision-specific ablations (\cref{sec:train-ablation}), and finally analyze cross-domain generalization (\cref{sec:train-generalization}). Additional training details and empirical results are deferred to the supplementary materials.





\subsection{Algorithm Comparison}
\label{sec:train-algo}

We evaluate three RL algorithms---GRPO, GSPO, and SAPO---across 16 Gym-V environments (12 single-turn and 4 multi-turn) under a unified experimental setup.
All algorithms share identical compute budgets and hyperparameter settings, differing only in their algorithm-specific update rules.

\begin{figure*}[tb]
  \centering
  \includegraphics[width=\linewidth]{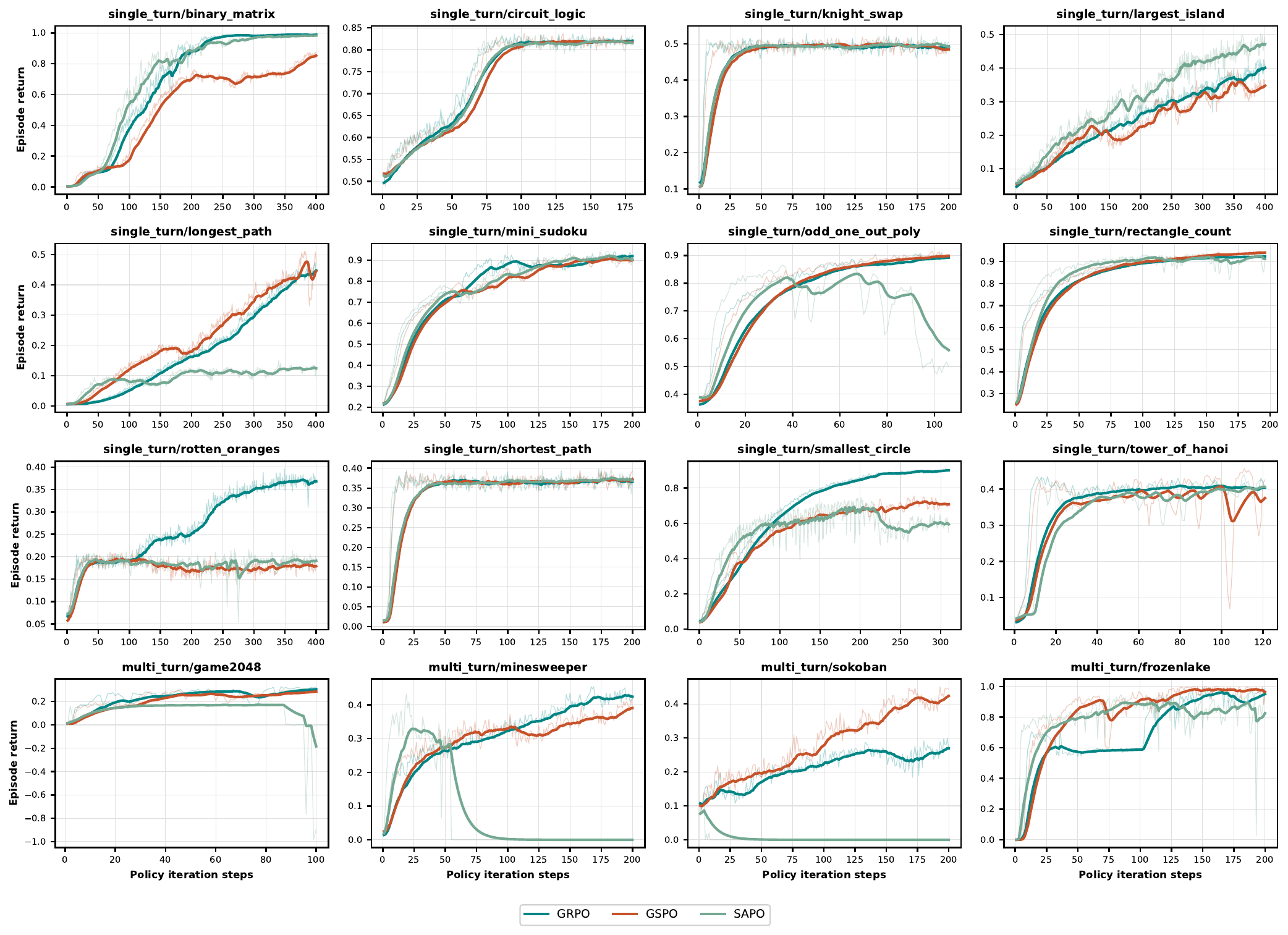}
  \caption{Training reward curves for GRPO, GSPO, and SAPO across 12 single-turn (rows~1--3) and 4 multi-turn (row~4) environments.
    Smoothed curves (bold) are overlaid on raw trajectories (translucent).
    All environments exhibit learnable reward signals; no single algorithm dominates uniformly.
    Multi-turn games show slower convergence and lower absolute returns, reflecting the compounding difficulty of sequential decision-making.
  }
  \label{fig:algo-curves}
\end{figure*}

Figure~\ref{fig:algo-curves} compares GRPO, GSPO, and SAPO for training \textbf{Qwen2.5-VL-7B} on 16 Gym-V environments (12 single-turn and 4 multi-turn) under identical scaffolding and compute. Across single-turn tasks, all three methods consistently improve performance, suggesting that the shared scaffolding and reward specification largely determine learnability, while algorithm choice mainly affects convergence speed and late-stage robustness. Notably, RL training can lift a 7B model performance beyond larger models with zero-shot evaluation setting on several environments: for example, the trained 7B reaches near-saturated performance on tasks like \texttt{binary\_matrix} and \texttt{circuit\_logic}, exceeding the zero-shot performance of Qwen2.5-VL-32B/72B (35.4\%/35.4\% and 37.5\%/62.5\%, respectively; . 

Multi-turn environments further accentuate update-stability differences: GSPO is the most consistent across \texttt{game2048}/\texttt{minesweeper}/\texttt{sokoban}/\texttt{frozenlake}, which we attribute to its sequence-level importance ratios and clipping being better matched to long-horizon rollouts than GRPO’s token-level ratios; in contrast, SAPO can collapse in harder multi-turn settings (e.g., \texttt{minesweeper}, \texttt{sokoban}), plausibly because its temperature-controlled off-policy gating becomes more sensitive under larger policy drift. Together, these results indicate that no single RL algorithm dominates across all settings: the best choice is scenario-dependent, with differences most visible in long-horizon multi-turn stability. In the following sections, we further dissect other training factors to understand what drives these outcomes.

\subsection{Diagnosing Multi-Turn and Vision-Centric Training}
\label{sec:train-ablation}

Multi-step vision-centric RL can be sensitive to seemingly small choices in how we present context, guidance, and observations.
We study three design keys to understand their impact on training.

\subsubsection{History Length in Multi-Turn Interaction.}
Long visual context can provide useful signals (e.g., past feedback and action effects), but may also introduce redundant or stale information that hurts performance. We compare three history settings: \emph{MDP} (current observation only), \emph{recent-3} (a sliding window of the last 2 observation--action pairs), and \emph{recent-5} (a sliding window of the last 4 observation--action pairs).

\noindent As shown in the top row of \Cref{fig:context-rules-ablation}, interaction history universally benefits multi-turn learning, but the degree of benefit correlates with environment complexity. The effect is strongest in environments with irreversible dynamics: in \texttt{Sokoban}, the MDP agent barely improves over the baseline, while context-equipped agents learn effectively---without observing recent action outcomes, the agent cannot identify which pushes led to dead states. As environments become more forgiving, context shifts from being \emph{necessary} for learning to being a \emph{catalyst}: in \texttt{Minesweeper} and \texttt{Game2048}, context accelerates convergence and improves early training, while in \texttt{FrozenLake}, all settings eventually reach near-optimal returns but context reduces variance and speeds up convergence.

Between the two window lengths, the 5-turn window is generally equal to or better than 3-turn: it achieves a higher final return in \texttt{Sokoban} and more stable training in \texttt{Game2048} (where the 3-turn window exhibits late-stage instability), while the two are comparable in simpler settings.

\subsubsection{Explicit Guidance vs.\ Autonomous Exploration.}
A key capability for agentic RL is whether an agent can discover environment constraints and valid actions from interaction, even when instructions are minimal. Recent work on rule induction from trajectories highlights the importance of this ability in multi-step settings~\citep{wang2025cogito}. We probe this in Gym-V by toggling whether the prompt includes the game rules.

\begin{figure*}[tb]
  \centering
  \includegraphics[width=\linewidth]{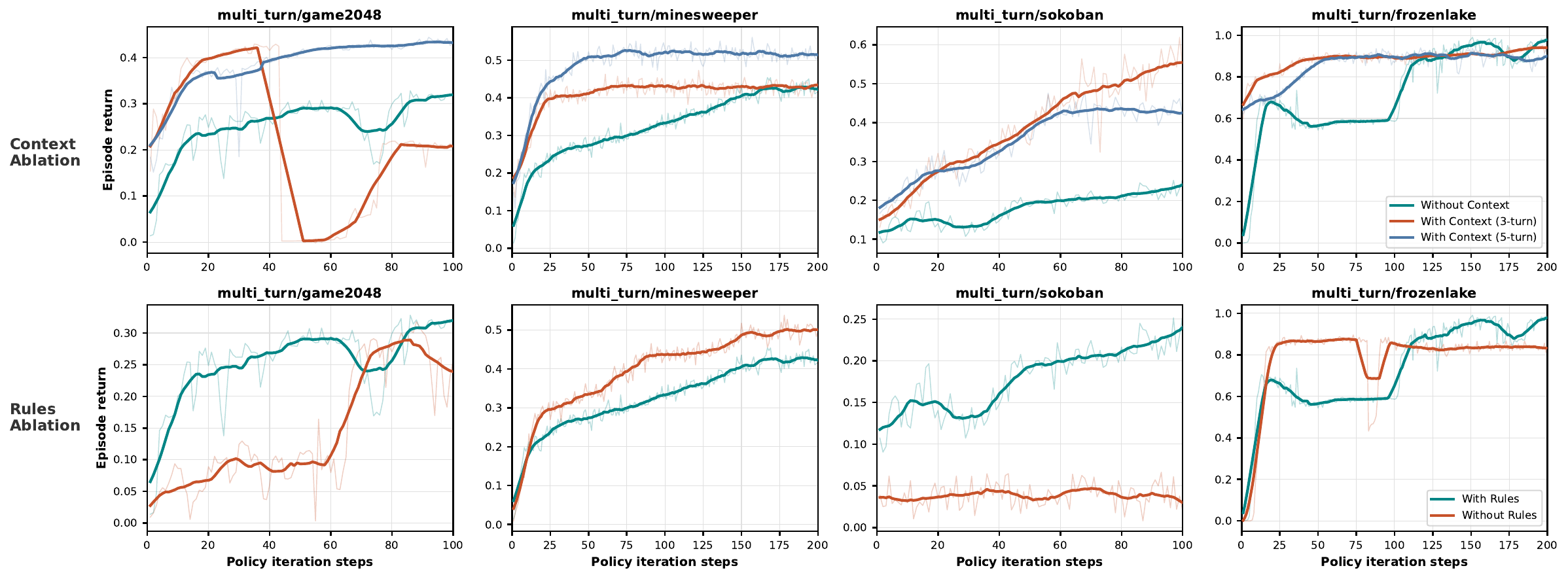}
  \caption{Training reward curves for context modeling (top row) and rules injection (bottom row) ablations on four multi-turn games.
    \textbf{Top}: \emph{with context} (3-turn, red; 5-turn, blue) vs.\ \emph{without context} (green).
    \textbf{Bottom}: \emph{with rules} (green) vs.\ \emph{without rules} (red).
    Smoothed curves (bold) are overlaid on raw trajectories (translucent).
  }
  \label{fig:context-rules-ablation}
\end{figure*}

\noindent As shown in Figure~\ref{fig:context-rules-ablation}, including game rules in the prompt generally improves learning, though the magnitude varies across environments. In \texttt{Sokoban}, training without rules largely fails to improve, whereas providing rules enables steady gains, suggesting that inferring valid actions and constraints purely from interaction is difficult under long-horizon, irreversible dynamics. We observe a similar advantage of rule prompting in \texttt{FrozenLake} and \texttt{2048}, where the \emph{with-rules} runs achieve higher returns. \texttt{Minesweeper} is an exception: removing rules yields higher reward, possibly because the game provides dense local feedback (revealed numbers) that supports rapid heuristic learning, while rule text can distract from the visual signal or bias overly cautious play under the shaped reward. Overall, rule prompting is beneficial in most cases, indicating that improving autonomous exploration and constraint discovery remains an important direction for future algorithmic development.

\subsubsection{Text Scaffolding for Visual Observations.}
Task representation can materially affect learning in visually interactive settings. To quantify the role of auxiliary text, we compare it on two single-turn tasks (\texttt{largest\_island}, \texttt{longest\_path}) and two multi-turn games (\texttt{minesweeper}, \texttt{frozenlake}) with two observation variants: \emph{image only} and \emph{image + caption}

\begin{figure*}[tb]
  \centering
  \includegraphics[width=\linewidth]{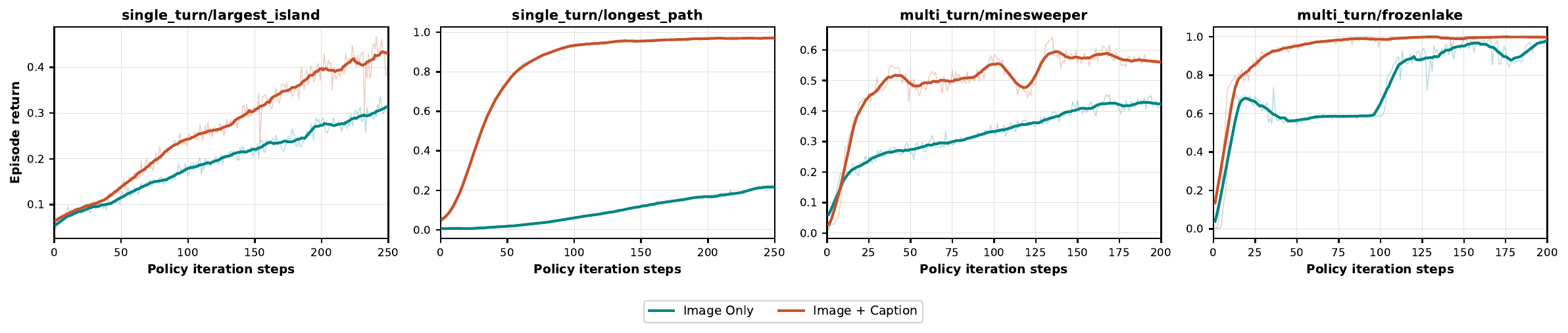}
  \caption{Training reward curves comparing \emph{image only} (green) vs.\ \emph{image + caption} (red) on two single-turn and two multi-turn environments.
    Adding textual captions yields a substantial and consistent improvement across all four environments.
    The benefit is most pronounced on perception-heavy tasks such as \texttt{longest\_path}, where caption narrows the visual grounding gap early in training, and on multi-turn games (\texttt{minesweeper}, \texttt{frozenlake}), where natural-language state descriptions help the agent maintain situational awareness across interaction steps.
  }
  \label{fig:caption-ablation}
\end{figure*}

As shown in \Cref{fig:caption-ablation}, captions consistently improve learning across all four environments. The effect is most pronounced on \texttt{longest\_path}: the image-only agent makes little progress, while the captioned agent rapidly reaches near-optimal return, indicating that performance is strongly limited by extracting the relevant state from raw pixels. Captions also yield higher returns on \texttt{largest\_island}, and substantially improve both learning speed and final performance in multi-turn settings (\texttt{minesweeper}, \texttt{frozenlake}). Overall, these results suggest that explicit linguistic grounding of the visual state can significantly ease optimization in vision-centric RL.

\subsection{Domain Generaltization}
\label{sec:train-generalization}

Gym-V supports both training and evaluation with the same environment interface, enabling controlled studies of cross-environment generalization.
We therefore ask whether RL training on one category transfers to unseen categories.
For each of the six single-turn categories, we train on all environments within that category except two held-out test environments, and evaluate the resulting model on the held-out test sets of all six categories.
This yields a $6{\times}6$ transfer matrix (\cref{tab:transfer-matrix} (a)). For Multi-turn setting, we training on one game and evaluate on four games, which yields a $4{\times}4$ transfer matrix (\cref{tab:transfer-matrix} (b)).

\begin{table*}[t]
  \caption{Cross-domain transfer matrices ($\times 100$) for \textbf{(a)}~single-turn categories and \textbf{(b)}~multi-turn games.
    Each row is a training source; each column an evaluation target.
    Subscripts show absolute change from zero-shot baseline.
    Diagonal entries (bold) denote in-domain performance.
  }
  \label{tab:transfer-matrix}
  \centering

  \begin{minipage}[t]{0.58\linewidth}
    \centering
    \textbf{(a)} Single-turn\\[3pt]
    \resizebox{\linewidth}{!}{%
      \setlength{\tabcolsep}{2.5pt}%
      \begin{tabular}{@{}l cccccc@{}}
        \toprule
        Train $\backslash$ Eval
          & Algo.
          & Cogn.
          & Geom.
          & Graph
          & Logic
          & Puzzle \\
        \midrule
        Baseline
          & 3.6 & 31.6 & 5.3 & 1.2 & 36.7 & 7.5 \\
        \midrule
        Algorithmic
          & \textbf{4.5}\,{\scriptsize$_{+0.9}$}
          & 47.1\,{\scriptsize$_{+15.5}$}
          & 7.4\,{\scriptsize$_{+2.1}$}
          & 1.6\,{\scriptsize$_{+0.5}$}
          & 45.8\,{\scriptsize$_{+9.2}$}
          & 8.1\,{\scriptsize$_{+0.5}$} \\
        Cognition
          & 5.7\,{\scriptsize$_{+2.2}$}
          & \textbf{48.3}\,{\scriptsize$_{+16.7}$}
          & 8.5\,{\scriptsize$_{+3.2}$}
          & 1.7\,{\scriptsize$_{+0.5}$}
          & 42.5\,{\scriptsize$_{+5.8}$}
          & 12.2\,{\scriptsize$_{+4.7}$} \\
        Geometry
          & 2.5\,{\scriptsize$_{-1.1}$}
          & 36.2\,{\scriptsize$_{+4.6}$}
          & \textbf{11.9}\,{\scriptsize$_{+6.6}$}
          & 0.8\,{\scriptsize$_{-0.3}$}
          & 42.6\,{\scriptsize$_{+5.9}$}
          & 4.1\,{\scriptsize$_{-3.4}$} \\
        Graphs
          & 3.7\,{\scriptsize$_{+0.1}$}
          & 40.3\,{\scriptsize$_{+8.7}$}
          & 8.1\,{\scriptsize$_{+2.8}$}
          & \textbf{1.3}\,{\scriptsize$_{+0.1}$}
          & 45.1\,{\scriptsize$_{+8.4}$}
          & 11.0\,{\scriptsize$_{+3.5}$} \\
        Logic
          & 5.8\,{\scriptsize$_{+2.3}$}
          & 42.6\,{\scriptsize$_{+11.0}$}
          & 7.2\,{\scriptsize$_{+1.9}$}
          & 1.1\,{\scriptsize$_{-0.1}$}
          & \textbf{45.6}\,{\scriptsize$_{+9.0}$}
          & 10.5\,{\scriptsize$_{+3.0}$} \\
        Puzzles
          & 5.4\,{\scriptsize$_{+1.9}$}
          & 40.1\,{\scriptsize$_{+8.5}$}
          & 5.1\,{\scriptsize$_{-0.3}$}
          & 2.6\,{\scriptsize$_{+1.4}$}
          & 46.5\,{\scriptsize$_{+9.9}$}
          & \textbf{14.4}\,{\scriptsize$_{+6.9}$} \\
        \bottomrule
      \end{tabular}%
    }
  \end{minipage}%
  \hfill
  \begin{minipage}[t]{0.40\linewidth}
    \centering
    \textbf{(b)} Multi-turn\\[3pt]
    \resizebox{\linewidth}{!}{%
      \renewcommand{\arraystretch}{1.35}%
      \setlength{\tabcolsep}{3pt}%
      \begin{tabular}{@{}l cccc@{}}
        \toprule
        Train $\backslash$ Eval
          & FrozenLake & Game2048 & Minesweeper & Sokoban \\
        \midrule
        Baseline
          & 0.3 & 0.2 & 1.1 & 5.5 \\
        \midrule
        FrozenLake
          & \textbf{97.7}\,{\scriptsize$_{+97.4}$}
          & $-$2.5\,{\scriptsize$_{-2.7}$}
          & 7.6\,{\scriptsize$_{+6.5}$}
          & 14.1\,{\scriptsize$_{+8.6}$} \\
        Game2048
          & 26.0\,{\scriptsize$_{+25.7}$}
          & \textbf{6.2}\,{\scriptsize$_{+6.0}$}
          & 4.8\,{\scriptsize$_{+3.7}$}
          & 12.5\,{\scriptsize$_{+7.0}$} \\
        Minesweeper
          & 45.3\,{\scriptsize$_{+45.0}$}
          & 1.2\,{\scriptsize$_{+1.0}$}
          & \textbf{42.4}\,{\scriptsize$_{+41.3}$}
          & 15.1\,{\scriptsize$_{+9.6}$} \\
        Sokoban
          & 35.7\,{\scriptsize$_{+35.4}$}
          & 1.1\,{\scriptsize$_{+0.9}$}
          & 6.2\,{\scriptsize$_{+5.1}$}
          & \textbf{31.8}\,{\scriptsize$_{+26.3}$} \\
        \bottomrule
      \end{tabular}%
    }
  \end{minipage}
\end{table*}

For single-turn environment: \noindent\textbf{broad curricula transfer, narrow ones hurt.}
We quantify a source domain's overall transfer capacity by its \emph{transfer breadth},
$B(s)=\sum_t \max(0,\Delta_{s\to t})$, i.e., the sum of positive improvements over the zero-shot baseline across targets. Cognition has the largest breadth ($33.1$), followed by Algorithmic ($28.7$) and Puzzles ($28.6$), whereas Geometry is lowest ($17.1$) and exhibits negative transfer. A plausible explanation is that Cognition and Puzzles cover diverse sub-skills (visual transformations, pattern completion, multi-step planning), encouraging reusable representations, while Geometry tasks are comparatively \emph{narrow}: many instances reduce to estimating a single numeric/geometric quantity from rendered point sets and shapes (e.g., \texttt{convex\_hull}, \texttt{smallest\_circle}, area/perimeter variants), which can be solved by direct ``read-off'' heuristics with limited procedural reuse. As a result, geometry-focused training may over-specialize to domain-specific shortcuts, yielding weaker transfer.

\noindent\textbf{Asymmetry implies a skill hierarchy.}
Transfer is not symmetric: Logic to Cognition yields $+11.0$ while Cognition to Logic yields $+5.8$, indicating that certain competencies act more like prerequisites than interchangeable skills.

\noindent\textbf{Negative transfer deep-dive.}
The strongest negative entries share a simple commonality: training can teach a ``default strategy'' that works in the source domain but is actively wrong in the target. Geometry$\to$Algorithmic ($-1.1$) likely reinforces a ``read-and-answer'' shortcut, while Algorithmic tasks require step-by-step simulation. FrozenLake$\to$Game2048 ($-2.7$) similarly promotes local, conservative navigation habits that conflict with 2048's global board reorganization under stochastic tile spawns, leading to confidently wrong behaviors and below-baseline performance.


For Multi-turn environment: \noindent\textbf{Multi-turn amplifies everything.}
\Cref{tab:transfer-matrix}(b) shows sparser and more fragile transfer among multi-turn games: in-domain training dominates, cross-game gains are limited (except that multiple sources substantially improve FrozenLake), and negative transfer can appear (e.g., FrozenLake to 2048). This aligns with the broader observation that multi-turn interaction compounds perception errors and policy drift across steps, making both algorithm stability and scaffolding choices (rules, captions, context) disproportionately important for generalization.

\section{Conclusion}
\label{sec:conclusion}

We presented Gym-V, a unified gym-style platform for vision-language models with a consistent API for single- and multi-turn training, multi-agent interaction, tool use, and generative evaluation, together with 179 procedurally seeded vision environments across 11 categories with difficulty presets and controllable distribution shifts. Using Gym-V as a testbed, we find that (i) Gym-V is challenging for most models and supports exploration of various RL training problems, (ii) scaffolding choices such as rules and image captions are critical—especially in multi-turn settings, and (iii) broad curricula transfer better while narrow training can cause negative transfer.


\bibliographystyle{plainnat}
\bibliography{main}

\newpage
\appendix

\section*{Overview of the Appendix}

This Appendix is organized as follows:
\begin{itemize}
    \item Section~\ref{sec:appendix-training} provides the training protocol details;
    \item Section~\ref{sec:appendix-context-norule} analyzes context as implicit rule discovery;
    \item Section~\ref{sec:appendix-temporal} discusses temporal environment analysis;
    \item Section~\ref{sec:appendix-examples} showcases per-environment examples;
    \item Section~\ref{sec:appendix-per-env} presents per-environment evaluation results;
    \item Section~\ref{sec:appendix-catalog} provides the complete environment catalog.
\end{itemize}

\section{Training Protocol}
\label{sec:appendix-training}

All training experiments use \textbf{Qwen2.5-VL-7B-Instruct} as the base model, fine-tuned with the verl framework~\citep{verl}.
We freeze the vision tower throughout training and use a learning rate of $1\times10^{-6}$.

\paragraph{Single-turn settings.}
Each training iteration samples a batch of 256 instances, each generating 8 rollouts for advantage estimation.
We apply a KL penalty with coefficient 0.01 to prevent excessive deviation from the reference policy.

\paragraph{Multi-turn settings.}
Each training iteration samples 256 complete trajectories, with up to 200 interaction turns per episode.
The KL penalty is disabled, as we observe it can overly constrain exploration in long-horizon settings where the policy must deviate more substantially from the reference to discover effective strategies.

\section{Context as Implicit Rule Discovery}
\label{sec:appendix-context-norule}

In \cref{sec:train-ablation}, we separately examined the effects of interaction history (context) and explicit rule prompting on multi-turn training. Here, we study their \emph{interaction}: can context compensate for the absence of game rules by enabling the agent to infer constraints from experience?

\begin{figure*}[tb]
  \centering
  \includegraphics[width=\linewidth]{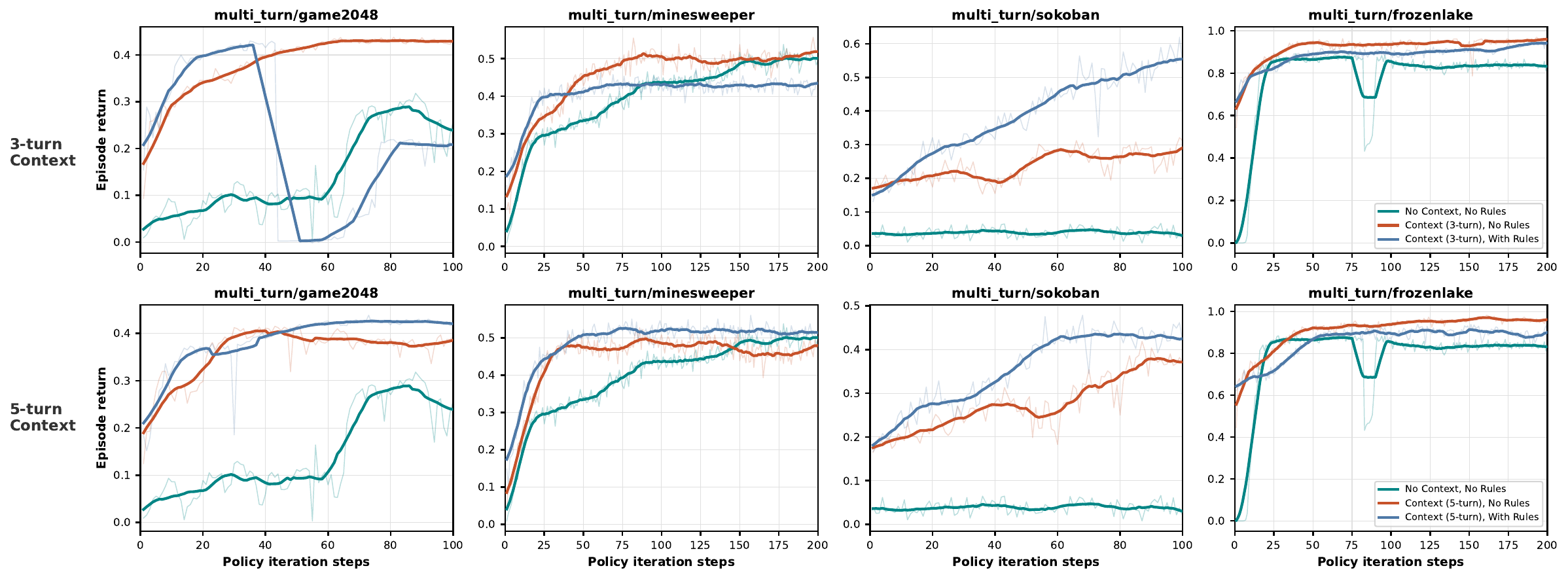}
  \caption{Training reward curves comparing three prompt conditions across four multi-turn games.
    \textbf{Top row}: 3-turn context window; \textbf{Bottom row}: 5-turn context window.
    Green: no context, no rules; Red: context only, no rules; Blue: context with rules.
    Providing interaction context without rules substantially outperforms the no-context baseline and, in several environments (\texttt{Game2048}, \texttt{FrozenLake}, \texttt{Minesweeper}), approaches or matches the with-rules setting, demonstrating that context enables implicit rule discovery through interaction experience.
  }
  \label{fig:context-norule-ablation}
\end{figure*}

\Cref{fig:context-norule-ablation} compares three conditions across four multi-turn games: (1)~\emph{No Context, No Rules}---the agent receives only the current observation; (2)~\emph{Context, No Rules}---the agent observes recent interaction history but receives no rule description; and (3)~\emph{Context, With Rules}---the agent has both history and explicit rules. We test with 3-turn and 5-turn context windows (top and bottom rows, respectively).

The results show that providing interaction context \emph{without} rules consistently and substantially outperforms the no-context baseline, and in several environments approaches the performance of the full context-with-rules setting. This effect is particularly pronounced in \texttt{Game2048} and \texttt{FrozenLake}, where the context-only agent nearly matches the rule-equipped agent, suggesting that a short interaction history provides sufficient signal for the model to infer the underlying game mechanics. In \texttt{Minesweeper}, the context-only agent even \emph{exceeds} the with-rules variant, consistent with our earlier observation that dense local feedback can substitute for explicit rule descriptions. \texttt{Sokoban} remains the most challenging: context helps, but a gap to the with-rules setting persists, reflecting the difficulty of discovering irreversible push constraints purely from interaction. Comparing the two context lengths, the 5-turn window generally provides a modest further improvement over 3-turn, with the largest gain in \texttt{Sokoban} where longer history helps diagnose dead-end states.

Overall, these results suggest that multi-turn context serves as a mechanism for \emph{implicit rule discovery}: even without explicit instructions, the agent can extract environment dynamics from its interaction history, partially closing the gap to the fully-informed setting.

\section{Temporal Environment Analysis}
\label{sec:appendix-temporal}

\begin{figure*}[tb]
  \centering
  \includegraphics[width=\linewidth]{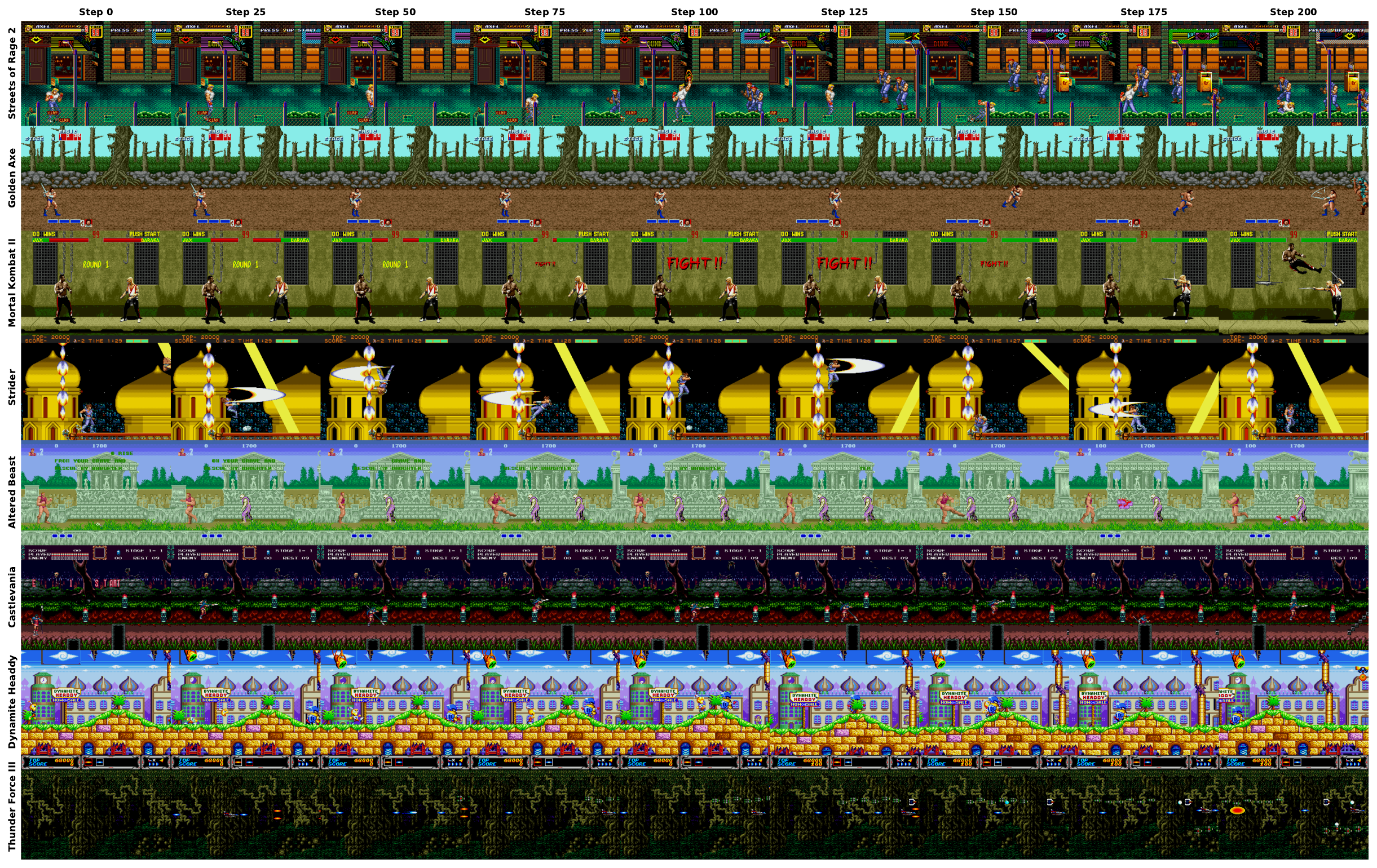}
  \caption{Sampled observations from 8 retro arcade environments over 200 VLM agent steps. Despite exhausting the full step budget, the game state barely progresses in any environment.}
  \label{fig:temporal-analysis}
\end{figure*}

Retro arcade games demand fine-grained, continuous control: meaningful progress (clearing a stage, defeating an opponent) requires sustained sequences of precise inputs over hundreds to thousands of consecutive frames. As shown in \Cref{fig:temporal-analysis}, within the allotted step budget, VLM agents cannot produce enough actions to trigger meaningful game events---characters remain near their starting positions after 200 steps, yielding near-zero reward across all temporal environments regardless of model capability. This category thus serves as a frontier benchmark for future systems capable of fine-grained temporal control.

\section{Per-Environment Examples}
\label{sec:appendix-examples}

We present representative environments from each of the 10 Gym-V categories below. For each environment, we show an example observation image; the caption briefly describes the task and expected output format. The complete list of all 202 environments, their configuration parameters, difficulty levels, and detailed implementation are available in the code repository.

\begin{figure}[tb]
  \centering
  \includegraphics[width=0.48\linewidth]{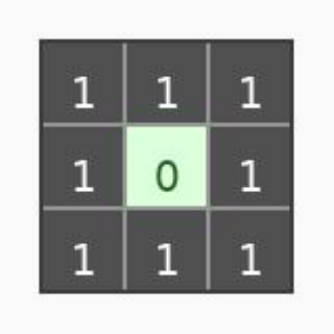}
  \hfill
  \includegraphics[width=0.48\linewidth]{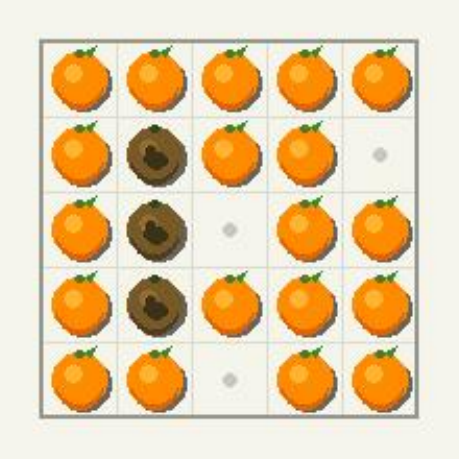}
  \caption{\textbf{Algorithmic.} \textit{Left:} BinaryMatrix --- the agent performs an algorithmic operation (e.g., counting connected components) on a rendered binary grid; output is a single integer or transformed grid. \textit{Right:} RottenOranges --- the agent determines the minimum time for all oranges to rot via BFS propagation on a grid; output is a single integer.}
  \label{fig:example-algorithmic}
\end{figure}


\begin{figure}[tb]
  \centering
  \includegraphics[width=0.48\linewidth]{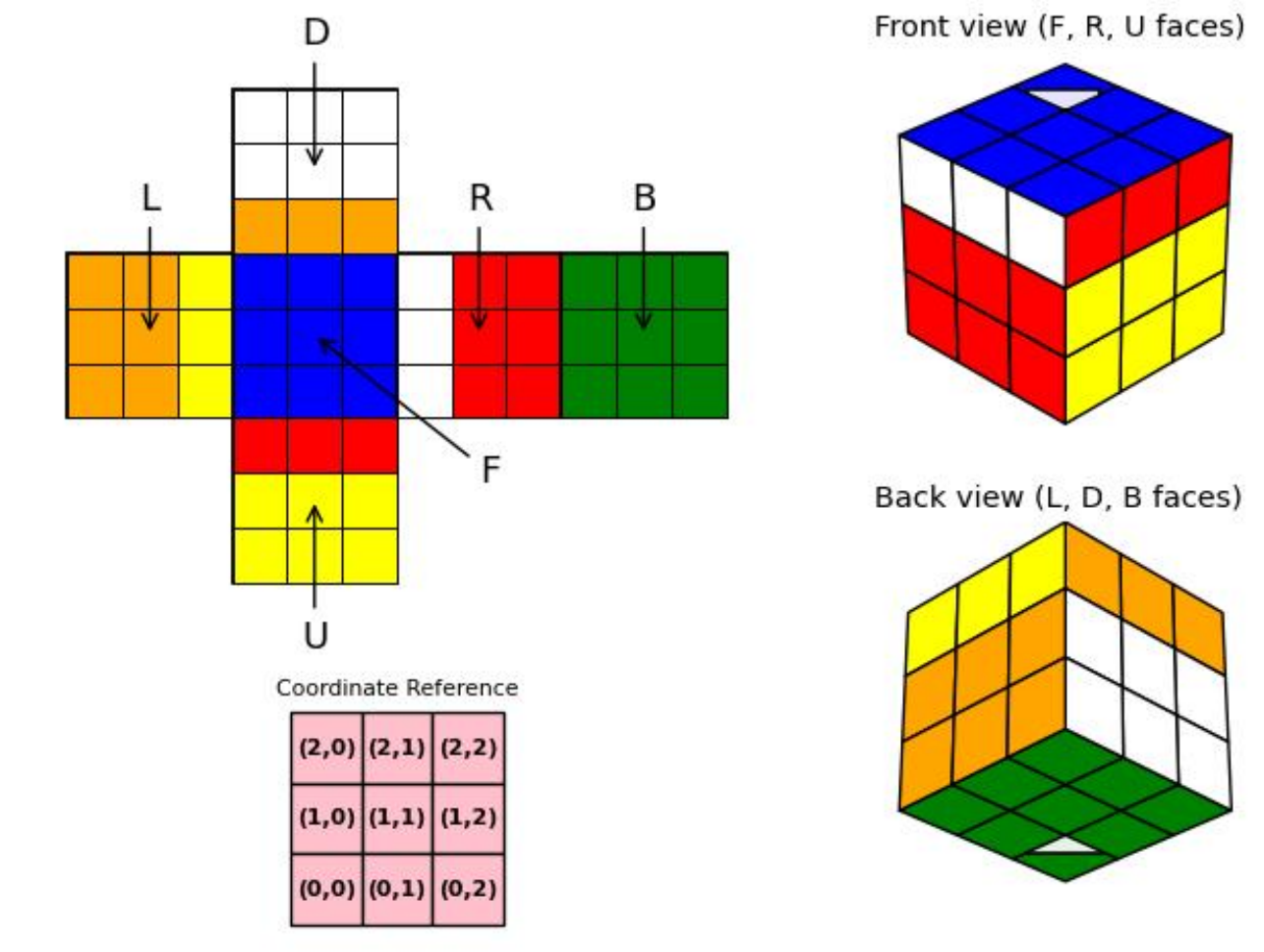}
  \hfill
  \includegraphics[width=0.48\linewidth]{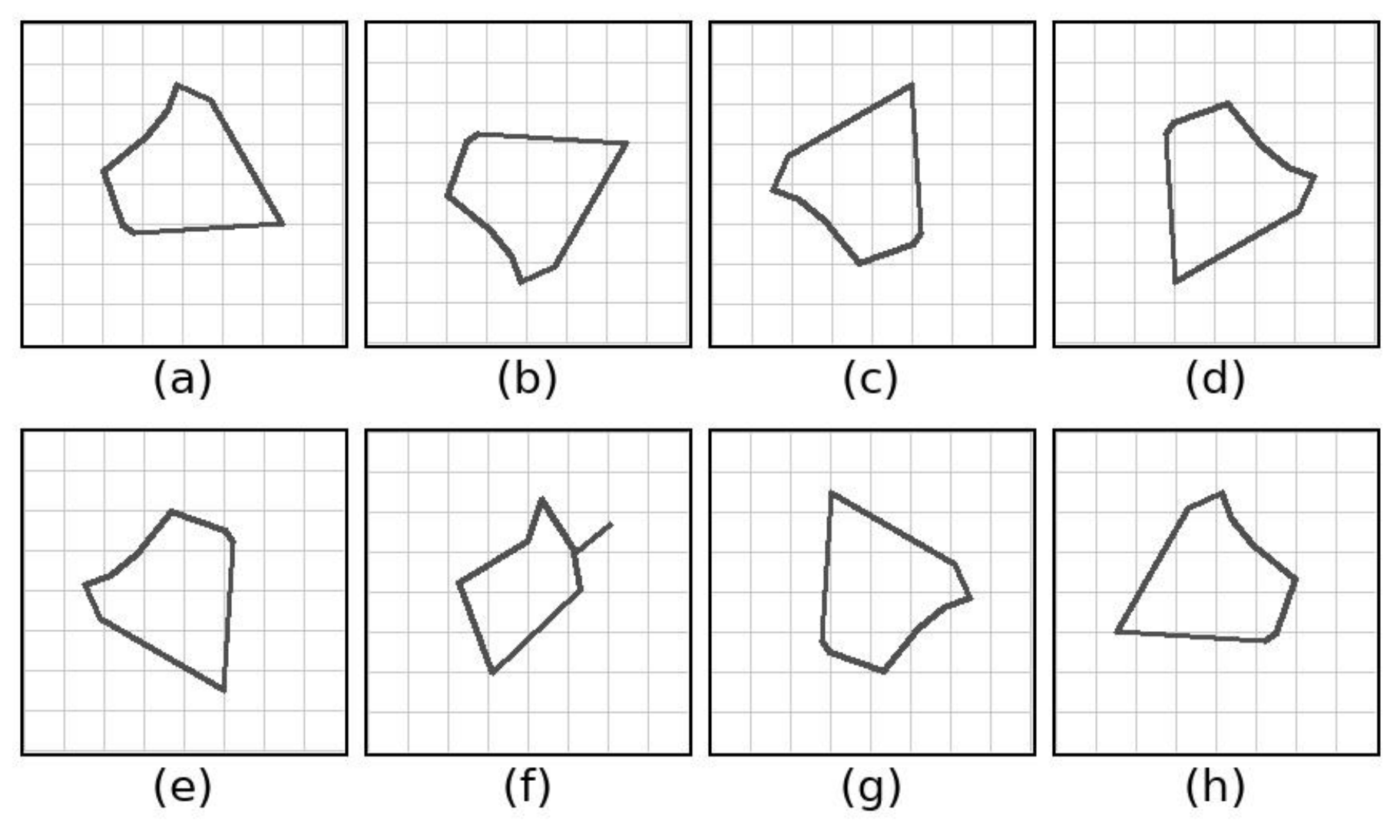}
  \caption{\textbf{Cognition.} \textit{Left:} RubiksCube-QA --- the agent observes a rendered 3D Rubik's Cube and answers questions about face colors after rotations. \textit{Right:} OddOneOutPoly --- the agent identifies which polygon differs from the rest based on a visual property (e.g., symmetry, number of sides).}
  \label{fig:example-cognition}
\end{figure}

\begin{figure}[tb]
  \centering
  \includegraphics[width=0.48\linewidth]{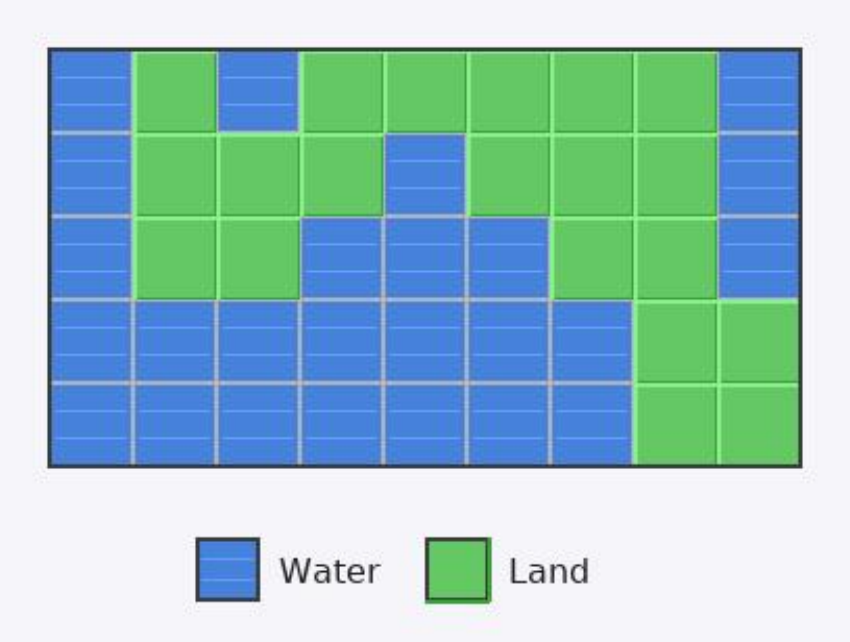}
  \hfill
  \includegraphics[width=0.48\linewidth]{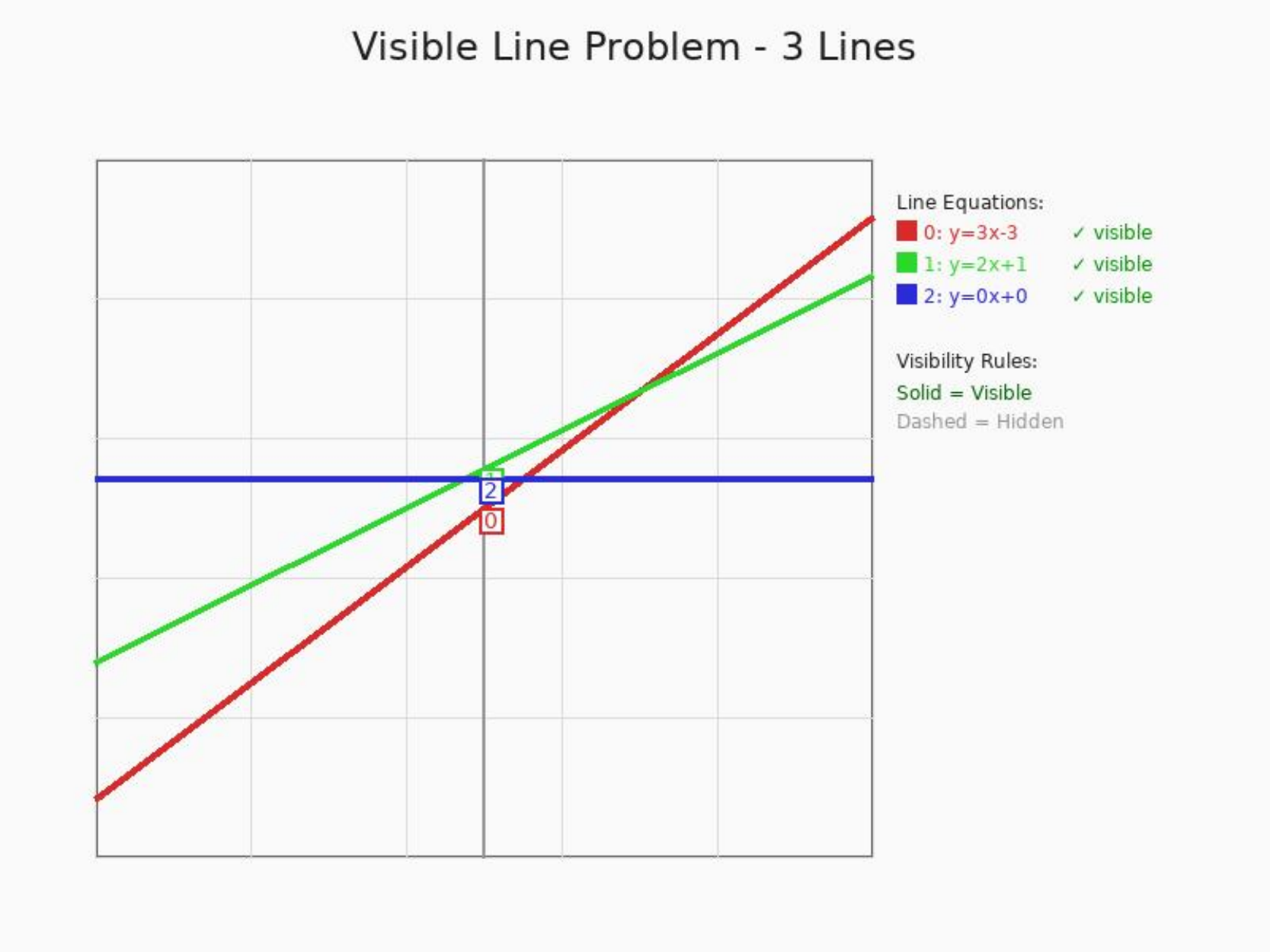}
  \caption{\textbf{Geometry.} \textit{Left:} LargestIsland --- the agent observes a binary grid (blue = water, green = land) and must find the maximum area of a 4-directionally connected island; output is a single integer. \textit{Right:} VisibleLine --- given $N$ lines $y=Ax+B$ plotted on a 2D plane, the agent identifies which lines are visible from $y=+\infty$ (i.e., lie on the upper envelope); output is the space-separated indices of visible lines.}
  \label{fig:example-geometry}
\end{figure}

\begin{figure}[tb]
  \centering
  \includegraphics[width=0.48\linewidth]{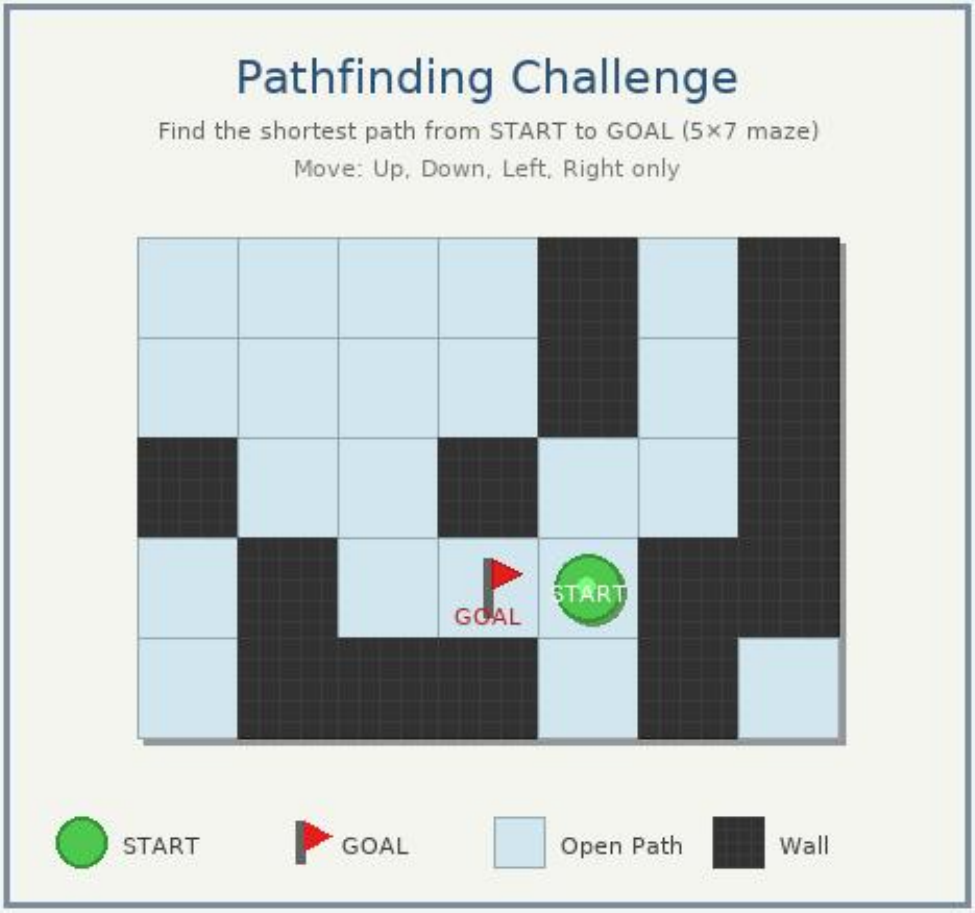}
  \hfill
  \includegraphics[width=0.48\linewidth]{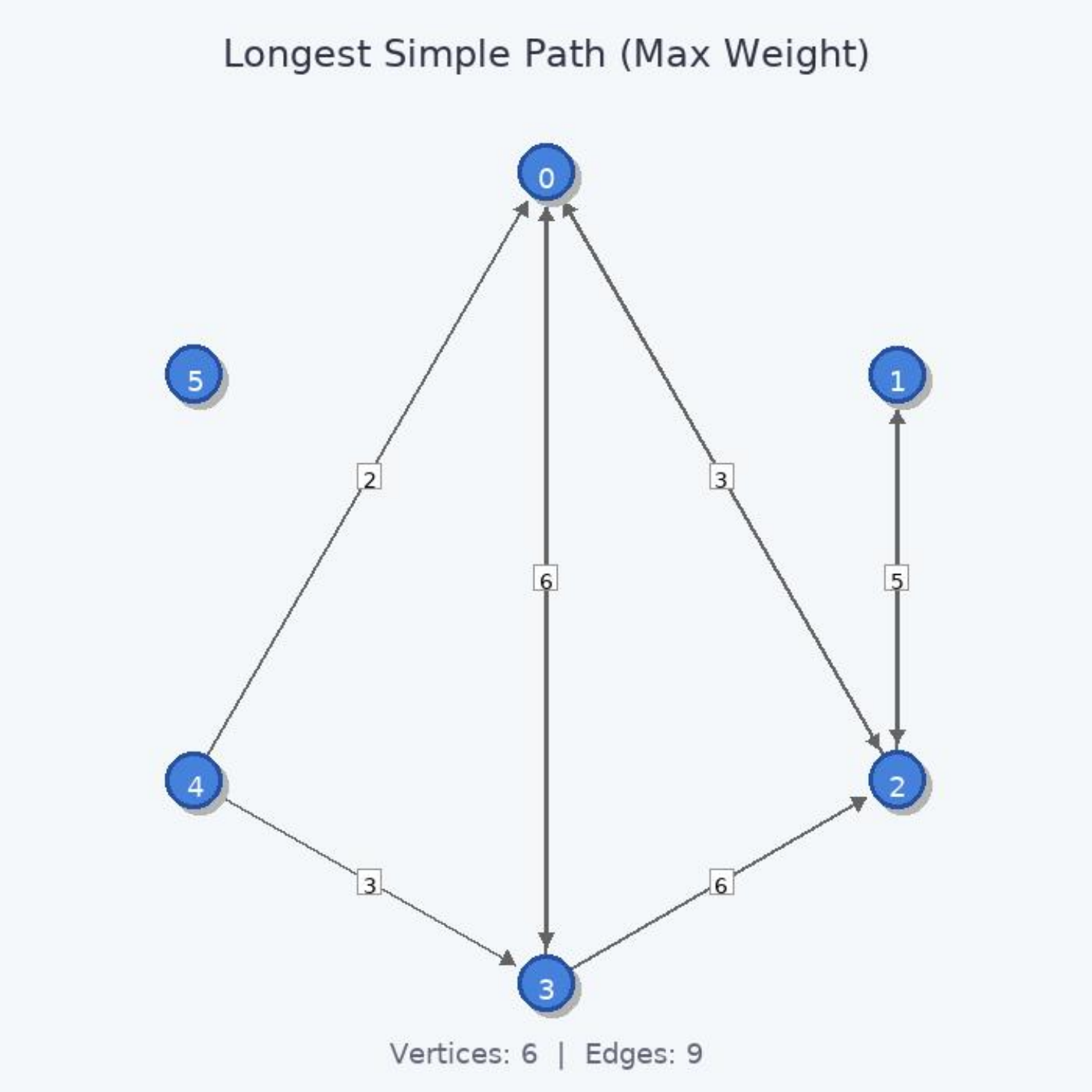}
  \caption{\textbf{Graphs.} \textit{Left:} ShortestPath --- the agent computes the shortest path between two nodes in a weighted graph; output is the node sequence. \textit{Right:} LongestPath --- the agent finds the longest simple path in a graph; output is the path length.}
  \label{fig:example-graphs}
\end{figure}

\begin{figure}[tb]
  \centering
  \includegraphics[width=0.48\linewidth]{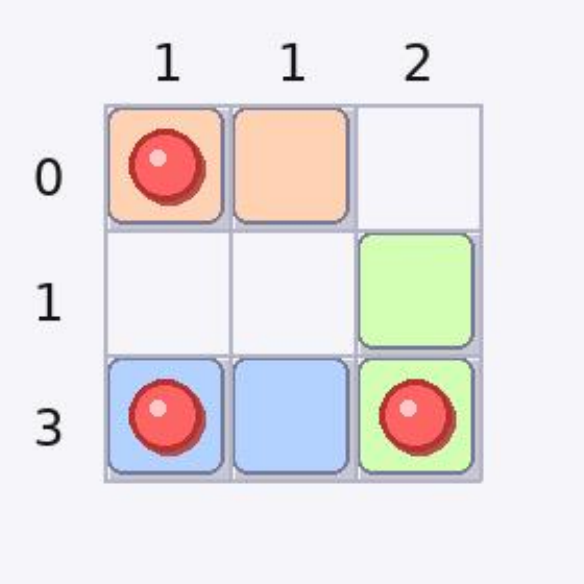}
  \hfill
  \includegraphics[width=0.48\linewidth]{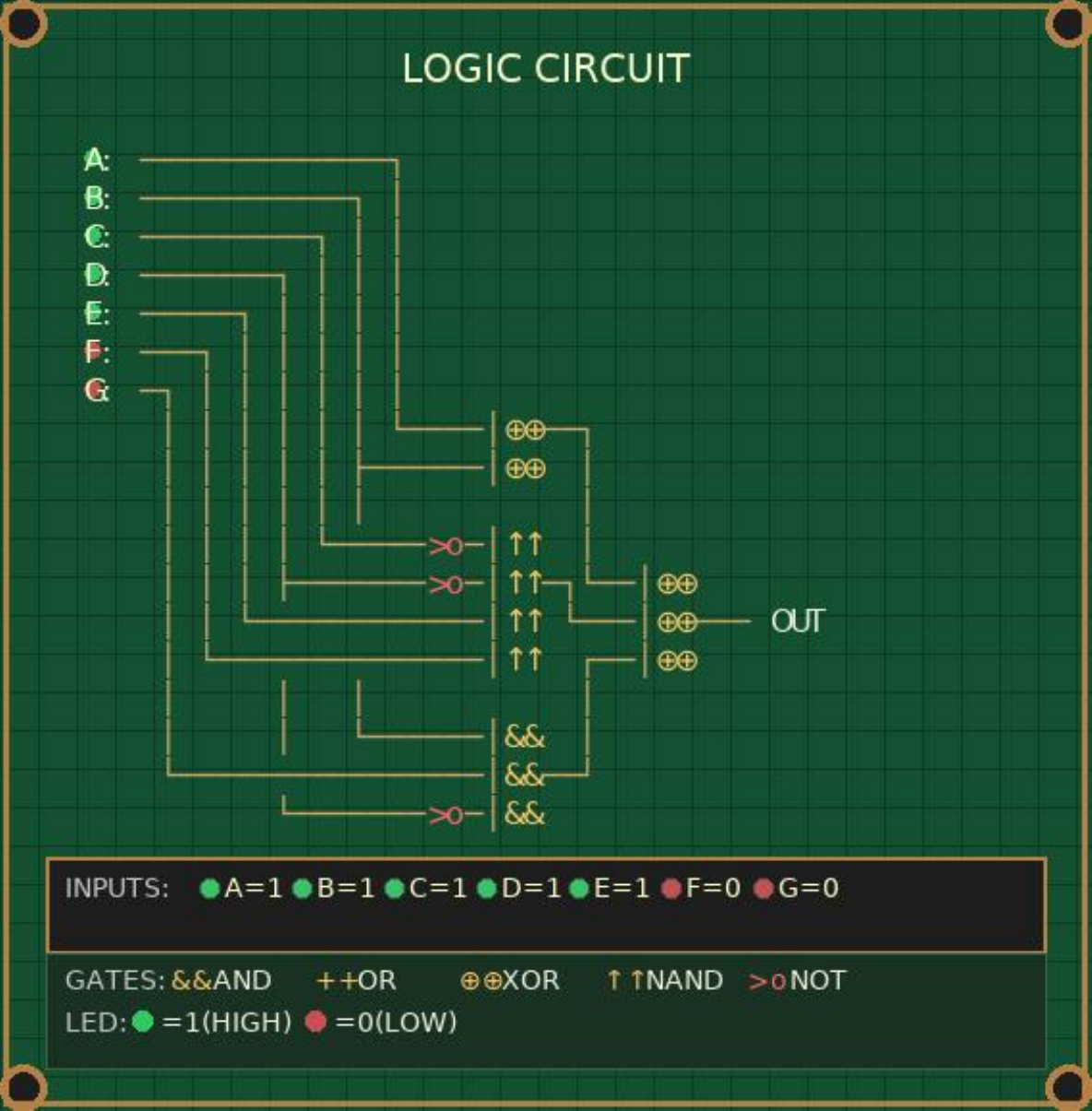}
  \caption{\textbf{Logic.} \textit{Left:} Thermometers --- the agent fills thermometer-shaped regions in a grid to satisfy row/column sum constraints. \textit{Right:} CircuitLogic --- the agent traces a Boolean logic circuit and determines the output given input values.}
  \label{fig:example-logic}
\end{figure}

\begin{figure}[tb]
  \centering
  \includegraphics[width=0.48\linewidth]{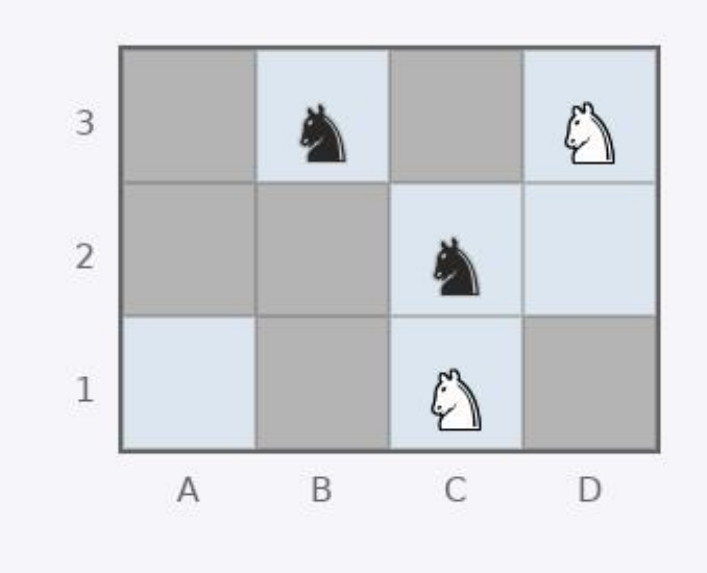}
  \hfill
  \includegraphics[width=0.48\linewidth]{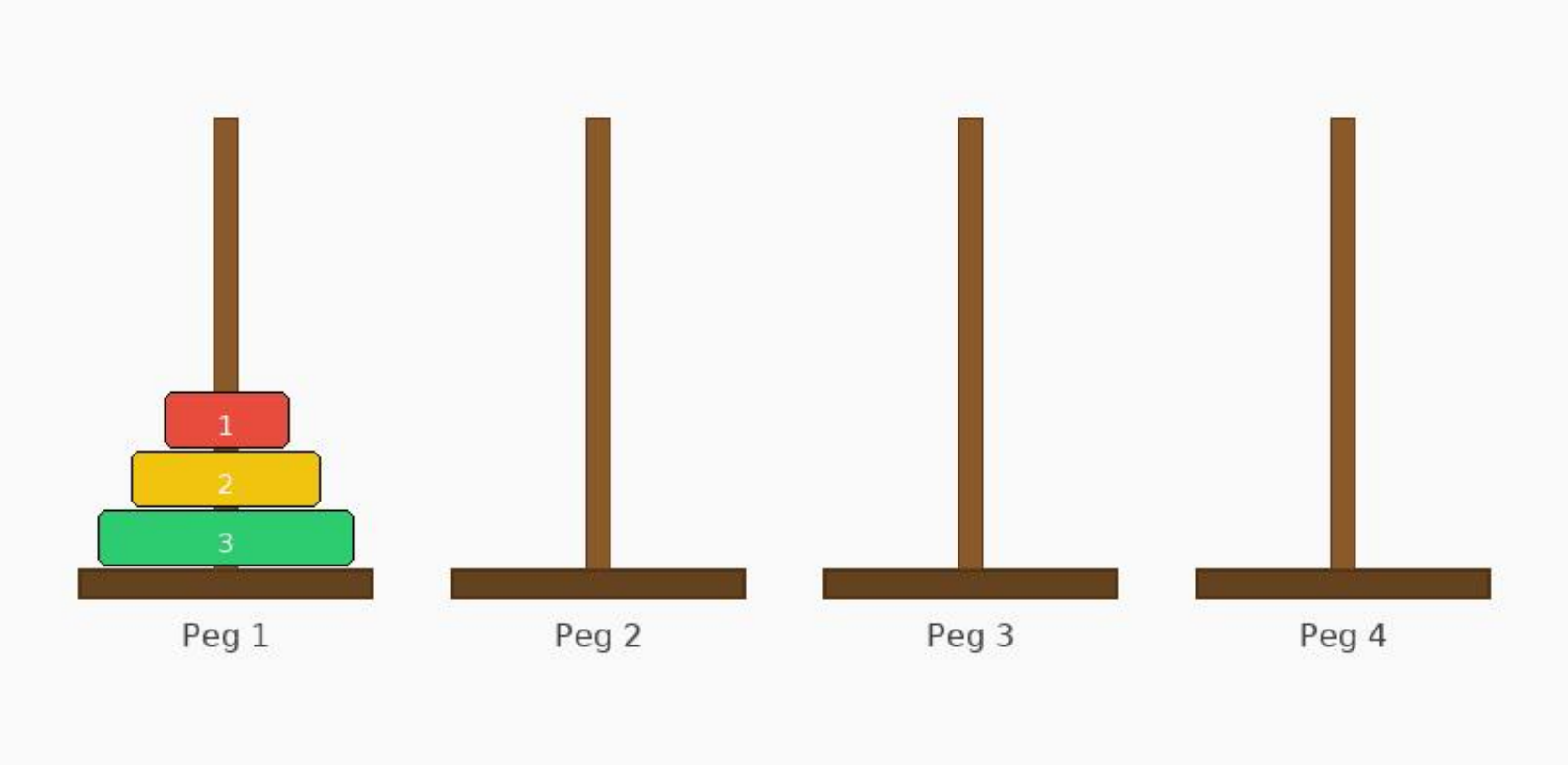}
  \caption{\textbf{Puzzles.} \textit{Left:} KnightSwap --- the agent swaps the positions of two sets of knights on a small chessboard using legal knight moves; output is the move sequence. \textit{Right:} TowerOfHanoi --- the agent produces the move sequence to transfer all discs to the target peg; output is a sequence of \texttt{(source, target)} moves.}
  \label{fig:example-puzzles}
\end{figure}

\begin{figure}[tb]
  \centering
  \includegraphics[width=0.48\linewidth]{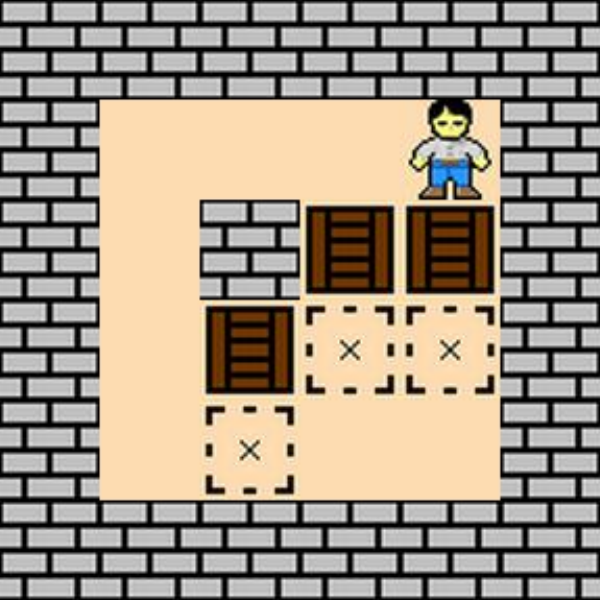}
  \hfill
  \includegraphics[width=0.48\linewidth]{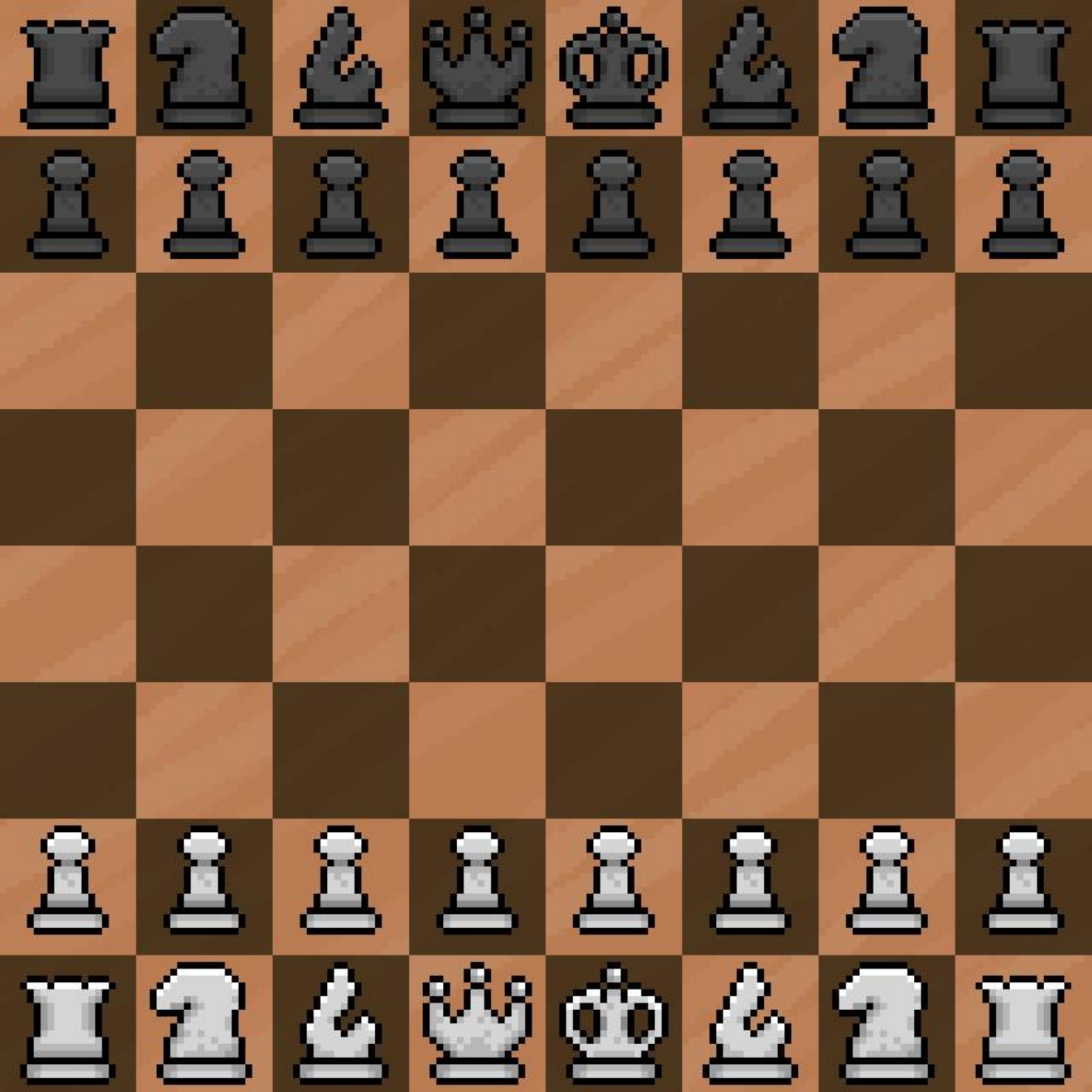}
  \caption{\textbf{Games.} \textit{Left:} Sokoban --- the agent pushes boxes onto goal positions by issuing directional commands (\texttt{up/down/left/right}) in a multi-turn grid puzzle. \textit{Right:} Chess --- the agent plays against a built-in opponent, outputting moves in UCI notation (e.g., \texttt{e2e4}) each turn.}
  \label{fig:example-games}
\end{figure}

\begin{figure}[tb]
  \centering
  \includegraphics[width=0.48\linewidth]{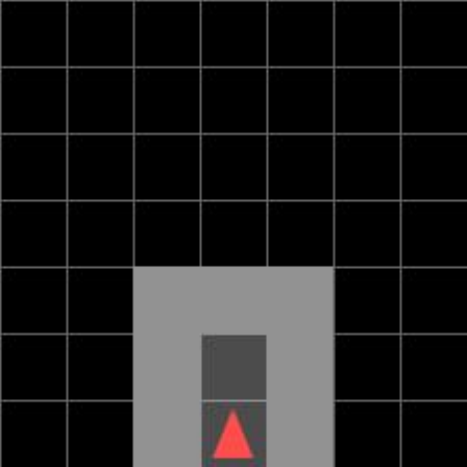}
  \hfill
  \includegraphics[width=0.48\linewidth]{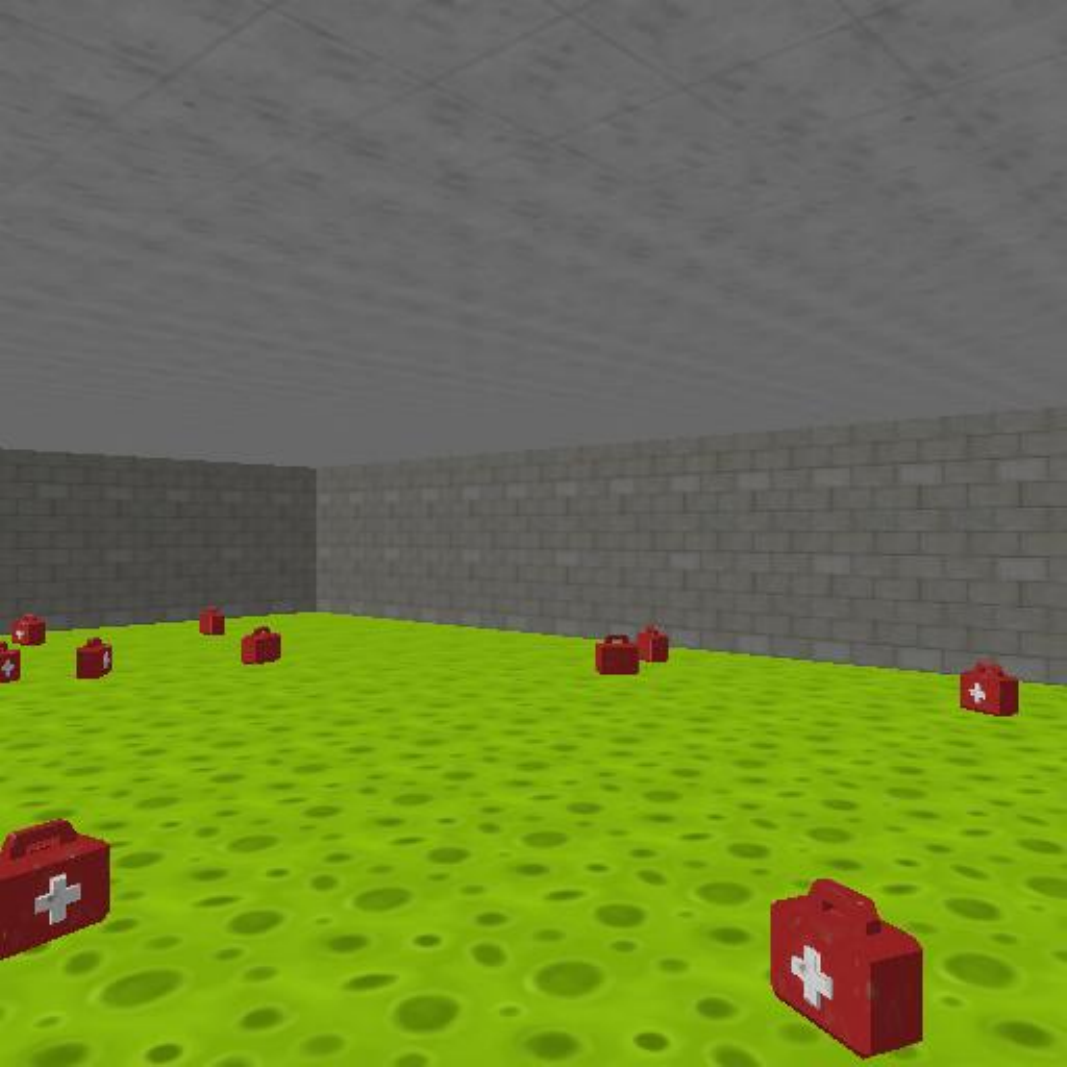}
  \caption{\textbf{Spatial.} \textit{Left:} DoorKey (MiniGrid, 2D) --- the agent navigates a top-down grid-world to find a key, unlock a door, and reach the goal. \textit{Right:} CollectHealth (MiniWorld, 3D) --- the agent navigates a first-person 3D room to collect health items; actions include \texttt{move\_forward} and \texttt{turn}.}
  \label{fig:example-spatial}
\end{figure}

\begin{figure}[tb]
  \centering
  \includegraphics[width=0.48\linewidth]{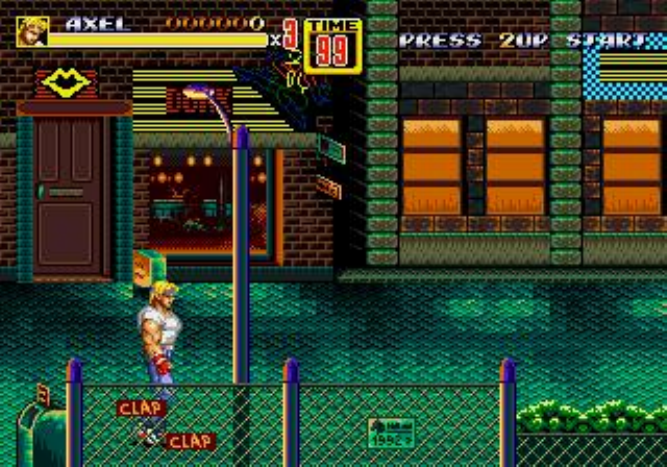}
  \hfill
  \includegraphics[width=0.48\linewidth]{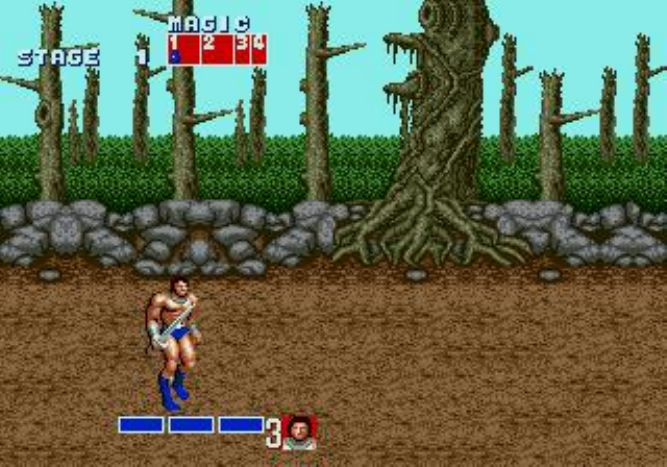}
  \caption{\textbf{Temporal.} \textit{Left:} StreetsOfRage2 --- a side-scrolling beat-em-up requiring continuous combat inputs. \textit{Right:} GoldenAxe --- a hack-and-slash arcade game.}
  \label{fig:example-temporal}
\end{figure}

\section{Per-Environment Evaluation Results}
\label{sec:appendix-per-env}

\begin{figure*}[p]
  \centering
  \includegraphics[width=\linewidth]{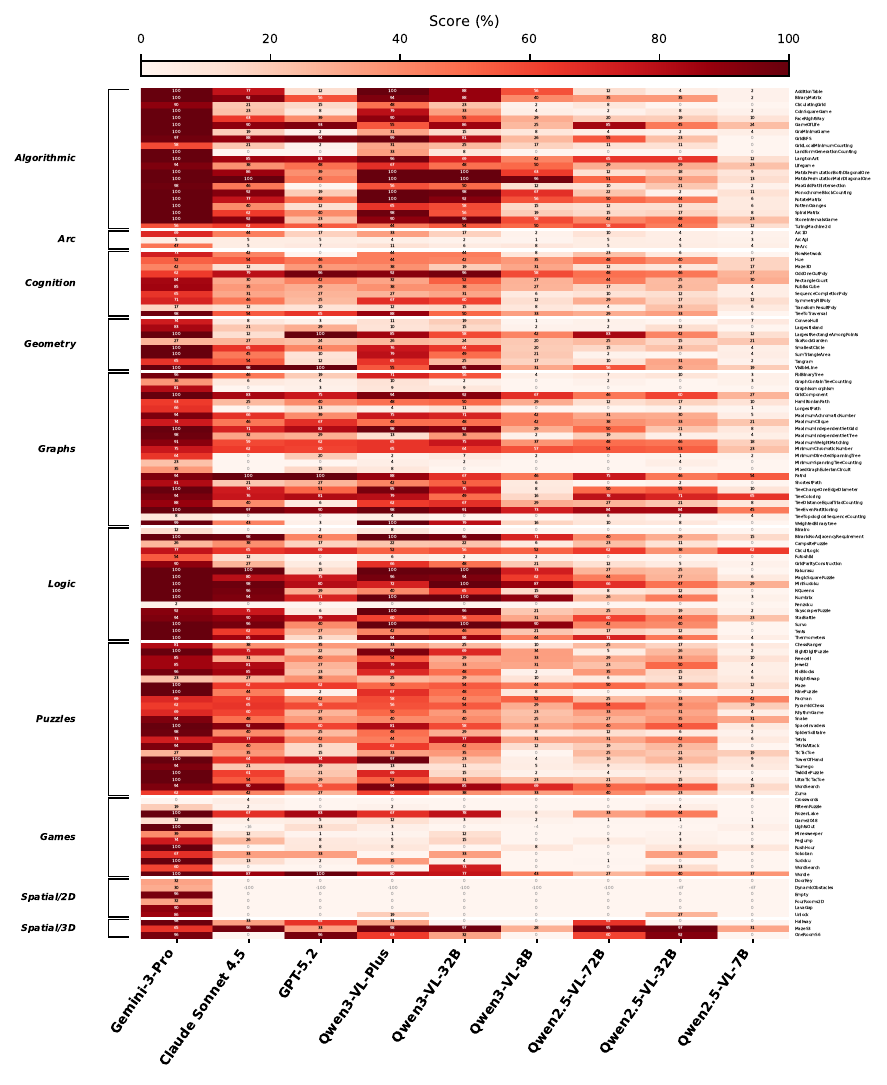}
  \caption{Per-environment zero-shot evaluation heatmap across all evaluated environments and 9 models. Rows are grouped by category; columns correspond to models ordered by overall performance. Scores are percentages ($\times 100$); darker shades indicate higher performance.}
  \label{fig:per-env-heatmap}
\end{figure*}

\Cref{fig:per-env-heatmap} presents the full per-environment breakdown of zero-shot mean@3 scores across all 9 evaluated models. Single-turn environments are scored by answer correctness, while multi-turn environments report normalized episodic return (negative values clipped to zero). All scores are shown as percentages ($\times 100$).

\section{Complete Environment Catalog}
\label{sec:appendix-catalog}

The following tables list all environments in Gym-V, organized by category and interaction mode. Each environment is accompanied by a brief task description derived from its implementation. Single-turn environments require the agent to produce a complete answer from a single observation, while multi-turn environments involve sequential decision-making over multiple interaction steps.

\onecolumn

\footnotesize
\renewcommand{\arraystretch}{1.25}

\begin{longtable}{>{\raggedright}p{0.43\textwidth} >{\raggedright\arraybackslash}p{0.53\textwidth}}
  \caption{Complete catalog of \textbf{single-turn} environments in Gym-V, grouped by category.}
  \label{tab:catalog-singleturn} \\
  \toprule
  \textbf{Environment} & \textbf{Description} \\
  \midrule
  \endfirsthead
  \toprule
  \textbf{Environment} & \textbf{Description} \\
  \midrule
  \endhead
  \midrule
  \multicolumn{2}{r}{\textit{Continued on next page}} \\
  \endfoot
  \bottomrule
  \endlastfoot

  \multicolumn{2}{@{}l}{\textit{\textbf{Algorithmic} (21 environments)}} \\
  \addlinespace[2pt]
  AdditionTable          & Find the correct base $N$ and digit mappings for a system where distinct letters represent digits, given addition equations. \\
  BinaryMatrix           & Compute the taxicab (Manhattan) distance to the nearest 0 for each cell in a binary matrix. \\
  CirculatingGrid        & Modify the minimum number of directional arrows in a toroidal grid so that every cell eventually loops back to itself. \\
  CoinSquareGame         & Determine the optimal strategy in a two-player coin-removal game where players take leftmost coins and sum their values. \\
  FaceRightWay           & Make all arrows in a binary array face right using the minimum number of fixed-window flip operations. \\
  GameOfLife             & Simulate Conway's Game of Life on a grid with wrapping boundaries. \\
  GraMinimaGame          & Determine optimal play in a two-player game where players select subsets, score the minimum, and remove chosen elements. \\
  GridBFS                & Compute shortest distances from each cell to the nearest target cell on a grid with impassable walls. \\
  GridLocalMinimumCounting & Count valid permutations of a grid that produce a target number of local minima among 8-connected neighbors. \\
  LandformGenerationCounting & Count distinct height sequences formed by valid permutations of terrain points under capacity constraints. \\
  LangtonAnt-QA          & Simulate and predict Langton's Ant cellular automaton on a grid with wrapping edges. \\
  Lifegame-QA            & Simulate Conway's Game of Life and answer questions about resulting cell states. \\
  MatrixPermBothDiag     & Find row and column permutations of a binary matrix so that both main and anti-diagonals contain only 1s. \\
  MatrixPermMainDiag     & Find row and column permutations of a binary matrix so that the main diagonal contains only 1s. \\
  MaxGridPathIntersection & Find the maximum total sum collected across $K$ top-left-to-bottom-right paths on a grid (first-visit collection). \\
  MonochromeBlockCounting & Determine the maximum number of monochrome layers in a tower and count distinct ways to achieve it. \\
  RotateMatrix           & Rotate a square matrix clockwise by a specified number of degrees. \\
  RottenOranges          & Determine the minimum time for all oranges to rot via BFS propagation on a grid. \\
  SpiralMatrix           & Extract matrix elements in clockwise spiral order starting from the top-left. \\
  StoneIntervalsGame     & Determine optimal play in a game where players collect entire piles adjacent to empty positions. \\
  TuringMachine2d-QA     & Simulate a 2D Turing machine following transition rules and predict the final tape state. \\
  \midrule
  \multicolumn{2}{@{}l}{\textit{\textbf{Arc} (3 environments)}} \\
  \addlinespace[2pt]
  Arc1D                  & Infer an abstract transformation rule from 1D input--output grid pairs and apply it to a test input. \\
  ArcAgi                 & Infer a transformation rule from 2D colored grid examples (ARC-AGI) and apply it to a test input. \\
  ReArc                  & Infer a transformation rule from procedurally generated 2D grid examples (ReARC) and apply it to a test input. \\
  \midrule
  \multicolumn{2}{@{}l}{\textit{\textbf{Cognition} (10 environments)}} \\
  \addlinespace[2pt]
  FlowNetwork            & Compute the maximum flow from source to sink in a visualized flow network with edge capacities. \\
  Hue-QA                 & Identify color gradient patterns in a hue puzzle where colors change gradually along rows or columns. \\
  Maze3D-QA              & Navigate a 3D maze by walking on cube tops and climbing ladders from start to goal. \\
  OddOneOutPoly          & Identify which of 8 polygons is not a rotation/reflection of the others. \\
  RectangleCount         & Count the number of rectangles (including overlapping ones) in an ASCII grid. \\
  RubiksCube-QA          & Determine Rubik's Cube face colors after a sequence of face rotations. \\
  SequenceCompletionPoly & Identify the next shape in a geometric sequence involving rotation and reflection. \\
  SymmetryFillPoly       & Select the shape that completes a 2$\times$2 grid to achieve vertical and horizontal mirror symmetry. \\
  TransformResultPoly    & Identify the result of a geometric transformation (rotation, reflection, or identity) applied to a shape. \\
  TreeToTraversal        & Extract preorder, inorder, and postorder traversal sequences from a visualized binary tree. \\
  \midrule
  \multicolumn{2}{@{}l}{\textit{\textbf{Geometry} (8 environments)}} \\
  \addlinespace[2pt]
  ConvexHull             & Find the convex hull of a set of 2D points (the smallest convex polygon containing all points). \\
  LargestIsland          & Find the largest 4-connected component of 1s in a binary matrix. \\
  LargestRectAmongPts    & Find 4 points forming a rectangle with maximum area (not necessarily axis-aligned). \\
  SkaRockGarden          & Determine which points to coordinate-swap to minimize the bounding rectangle perimeter. \\
  SmallestCircle         & Find the minimum enclosing circle for a set of 2D points. \\
  SumTriangleArea        & Compute the total area of all triangles formed by any three distinct points in a set. \\
  Tangram-QA             & Identify adjacency relationships among numbered tangram pieces, where some pieces are removed. \\
  VisibleLine            & Determine which lines ($y=Ax+B$) are visible from $y=+\infty$ (lie on the upper envelope). \\
  \midrule
  \multicolumn{2}{@{}l}{\textit{\textbf{Graphs} (23 environments)}} \\
  \addlinespace[2pt]
  FbiBinaryTree          & Classify binary string segments as F/B/I and output the postorder traversal of the resulting FBI tree. \\
  GraphContainTreeCount  & Count vertex bijections from a tree to a graph that preserve all tree edges. \\
  GraphIsomorphism       & Find an edge-preserving bijection between two undirected graphs. \\
  GridComponent          & Compute the largest connected component of 1s in a binary grid (4-connected). \\
  HamiltonianPath        & Find a minimum-weight path visiting every vertex in a directed weighted graph. \\
  LongestPath            & Find the maximum-weight simple path in a directed weighted graph. \\
  MaxAchromaticNumber    & Find a proper coloring using the maximum number of colors where every color pair shares an edge. \\
  MaximumClique          & Find the largest fully connected subset of vertices in an undirected graph. \\
  MaxIndepSetGrid        & Select non-adjacent cells in a matrix to maximize the sum of selected values. \\
  MaxIndepSetTree        & Select non-adjacent vertices in a weighted tree to maximize total weight. \\
  MaxWeightMatching      & Find a maximum-weight set of edges sharing no vertices in an undirected weighted graph. \\
  MinChromaticNumber     & Color graph vertices with the minimum number of colors so no two adjacent vertices share a color. \\
  MinDirSpanningTree     & Find a minimum-weight spanning arborescence rooted at a specified vertex in a directed graph. \\
  MinSpanTreeCounting    & Count the number of distinct minimum spanning trees modulo a given number. \\
  MixedGraphEulerian     & Orient undirected edges and find an Eulerian circuit in a mixed graph. \\
  Patrol                 & Find the minimum edges traversed in a round trip visiting all original tree edges and $K$ added edges. \\
  ShortestPath           & Find the shortest path between two nodes in a grid or graph with walls. \\
  TreeChangeEdgeDiam     & Remove and add one edge in a tree to minimize or maximize the resulting diameter. \\
  TreeColoring           & Select $K$ vertices to maximize total pairwise distance among colored and uncolored vertices. \\
  TreeDistEqualTriad     & Count three-vertex sets with all pairwise distances equal in a tree. \\
  TreeEvenPartition      & Partition a tree into connected subgraphs of exactly $K$ vertices each. \\
  TreeTopoSeqCounting    & Count permutations satisfying ordering constraints corresponding to a tree structure. \\
  WeightedBinarytree     & Construct a binary tree with fixed in-order traversal that maximizes a recursive score function. \\
  \midrule
  \multicolumn{2}{@{}l}{\textit{\textbf{Logic} (17 environments)}} \\
  \addlinespace[2pt]
  Binairo                & Fill a grid with black/white circles so each row and column has equal counts, with no three consecutive same-color. \\
  BinarioNoAdjacency     & Fill a grid with 0/1 so each row and column has exactly half of each (no adjacency constraint). \\
  CampsitePuzzle         & Fill empty cells with 0/1 satisfying adjacency and row/column count constraints. \\
  CircuitLogic           & Evaluate a randomly generated Boolean logic circuit given input values. \\
  Futoshiki              & Fill a grid with 1 to $N$ so each row/column has unique values, respecting inequality constraints. \\
  GridParityConstruction & Construct a binary matrix whose XOR-parity matrix matches a given target. \\
  Kakurasu               & Place 1s in a grid so the weighted sum of each row and column matches given clues. \\
  MagicSquarePuzzle      & Complete a grid to form a magic square where rows, columns, and diagonals share a common sum. \\
  MiniSudoku             & Solve a 4$\times$4 Sudoku where each row, column, and 2$\times$2 box contains 1--4 exactly once. \\
  NQueens                & Place $N$ queens on an $N\!\times\!N$ board so no two share a row, column, or diagonal. \\
  Numbrix                & Fill a matrix with consecutive integers so each is adjacent (horizontally/vertically) to its successor. \\
  Renzoku                & Fill a Latin square where dots indicate consecutive neighbors and no dot means non-consecutive. \\
  SkyscraperPuzzle       & Fill a grid with unique heights per row/column; edge clues specify visible building counts. \\
  StarBattle             & Place exactly $K$ stars per row, column, and region, with no two stars adjacent (including diagonally). \\
  Survo                  & Fill a grid so the last element of each row/column equals the sum of the remaining elements. \\
  Tents                  & Place tents adjacent to trees satisfying row/column counts; no two tents may be adjacent. \\
  Thermometers           & Fill thermometer shapes continuously from the bulb, satisfying row/column filled-cell counts. \\
  \midrule
  \multicolumn{2}{@{}l}{\textit{\textbf{Puzzles} (23 environments)}} \\
  \addlinespace[2pt]
  ChessRanger-QA         & Remove chess pieces via capture-only moves until a single piece remains on the board. \\
  EightDigitPuzzle       & Solve the 8-puzzle by sliding numbered tiles into the empty space to reach the goal configuration. \\
  Freecell-QA            & Analyze a FreeCell solitaire layout and determine valid moves or winning strategies. \\
  Jewel2-QA              & Eliminate grid elements by forming lines of three or more identical items. \\
  KloBlocks              & Maximize the longest contiguous subarray with all elements $\geq K$ by redistributing values. \\
  KnightSwap             & Swap positions of white and black knights on a grid using valid knight moves. \\
  Maze-QA                & Navigate around blue obstacles to reach the green goal from the red starting position. \\
  NinePuzzle             & Transform a digit grid into a target via cyclic row/column shifts with bounded magnitudes. \\
  Pacman-QA              & Analyze a Pac-Man game state and determine optimal moves to eat beans while avoiding ghosts. \\
  PyramidChess-QA        & Solve a chess variant on a pyramid-shaped board with level-based support constraints. \\
  RhythmGame-QA          & Score points by clicking falling blocks at the correct row, handling special block types. \\
  Snake-QA               & Navigate the snake to eat food while avoiding collisions with its own body. \\
  SpaceInvaders-QA       & Analyze a Space Invaders game state with enemies arranged in a labeled grid. \\
  SpiderSolitaire-QA     & Arrange cards in King-to-Ace sequences and move completed sequences to foundation piles. \\
  Tetris-QA              & Analyze a Tetris board state and determine optimal piece placement. \\
  TetrisAttack           & Remove all paired integers from an array using minimum adjacent swaps with auto-removal. \\
  TicTacToe-QA           & Analyze a TicTacToe board and determine the optimal next move or game outcome. \\
  TowerOfHanoi           & Produce the move sequence to transfer all discs to the target peg obeying size constraints. \\
  Tsumego                & Find the key move for Black to capture stones in a Go life-and-death problem. \\
  TwiddlePuzzle          & Rotate $K\!\times\!K$ subgrids by 90\textdegree{} counterclockwise to match a destination grid. \\
  UltraTicTacToe-QA      & Play Ultimate TicTacToe across 9 sub-boards; win by claiming three sub-boards in a row. \\
  WordSearch-QA          & Find hidden words in a letter grid placed in any of 8 directions. \\
  Zuma-QA                & Shoot colored marbles to form matching groups and clear the entire track. \\
\end{longtable}

\begin{longtable}{>{\raggedright}p{0.43\textwidth} >{\raggedright\arraybackslash}p{0.53\textwidth}}
  \caption{Complete catalog of \textbf{multi-turn} environments in Gym-V, grouped by category.}
  \label{tab:catalog-multiturn} \\
  \toprule
  \textbf{Environment} & \textbf{Description} \\
  \midrule
  \endfirsthead
  \toprule
  \textbf{Environment} & \textbf{Description} \\
  \midrule
  \endhead
  \midrule
  \multicolumn{2}{r}{\textit{Continued on next page}} \\
  \endfoot
  \bottomrule
  \endlastfoot

  \multicolumn{2}{@{}l}{\textit{\textbf{Games} (31 environments)}} \\
  \addlinespace[2pt]
  Alquerque              & Capture all opponent pieces or block their moves on a 5$\times$5 connected board. \\
  Breakthrough           & Advance pieces to the opponent's home row; move forward or capture diagonally. \\
  Chess                  & Play standard chess against an opponent, outputting moves in UCI notation. \\
  ConnectFour            & Drop pieces into a 7-column, 6-row grid; connect four in a line to win. \\
  ConnectFourMultiAgent  & Two-agent Connect Four where both sides are controlled by separate models. \\
  Crosswords             & Fill in a crossword grid one letter at a time given clue constraints. \\
  Crusade                & All pieces move like chess knights; maximize capture score over 40 moves. \\
  FifteenPuzzle          & Arrange tiles 1--15 in order by sliding them into the empty space. \\
  FrozenLake             & Navigate from start to goal on a frozen grid without falling into holes. \\
  Game2048               & Slide and combine numbered tiles on a grid to reach the target value. \\
  GinRummy               & Draw and discard cards to form melds and minimize deadwood; knock or go Gin. \\
  Go                     & Place stones to surround territory and capture opponent's stones on a Go board. \\
  LeducHoldem            & Play simplified poker with 6 cards, one private card, and one community card. \\
  LightsOut              & Toggle lights and their neighbors to turn all lights off. \\
  LinesOfAction          & Connect all pieces into a single 8-neighbor-connected group on an 8$\times$8 board. \\
  Minesweeper            & Reveal all safe cells without triggering mines on a grid. \\
  Nim                    & Remove objects from piles; the player taking the last object wins. \\
  Othello                & Place pieces to flank and flip opponent's pieces; have the most pieces at game end. \\
  PegJump                & Jump pegs over adjacent pegs into empty holes; finish with exactly one peg. \\
  RushHour               & Slide cars on a grid to free the red car and drive it out the right edge. \\
  SimpleTak              & Place stones to form a continuous path connecting two opposite board edges. \\
  Sokoban                & Push all boxes onto target positions without pulling or pushing through walls. \\
  Sudoku                 & Fill empty cells so each row, column, and 3$\times$3 box contains 1--9 exactly once. \\
  TexasHoldem            & Play Texas Hold'em poker with limit betting rules. \\
  TexasHoldemNoLimit     & Play No-Limit Texas Hold'em with unrestricted bet sizing. \\
  TicTacToe              & Place marks on a 3$\times$3 grid; three in a row wins. \\
  TowerOfHanoiMultiTurn  & Move disks between pegs one at a time, never placing a larger disk on a smaller one. \\
  UltimateTicTacToe      & Win three mini-boards in a row across a 3$\times$3 grid of TicTacToe sub-boards. \\
  WildTicTacToe          & Place either X or O in any empty cell; first to get three of the same mark in a row wins. \\
  WordSearch             & Find and highlight hidden words on a letter grid over multiple turns. \\
  Wordle                 & Guess a secret word in limited attempts using color-coded positional feedback. \\
  \midrule
  \multicolumn{2}{@{}l}{\textit{\textbf{Spatial} (30 environments)}} \\
  \addlinespace[2pt]
  \multicolumn{2}{@{}l}{\quad \textit{2D (MiniGrid)}} \\
  \addlinespace[1pt]
  DoorKey                & Find a key, unlock a door, and reach the green goal in a top-down grid-world. \\
  DynamicObstacles       & Reach the goal while avoiding moving obstacles; collision ends the episode. \\
  Empty                  & Navigate an empty grid to reach the green goal square. \\
  FourRooms2D            & Navigate through four rooms connected by wall gaps to reach the goal. \\
  LavaGap                & Cross a lava gap to reach the goal; stepping on lava ends the episode. \\
  MultiRoom              & Navigate through multiple connected rooms to reach the goal. \\
  Unlock                 & Find a key to unlock a door and reach the goal in a simple room. \\
  \addlinespace[3pt]
  \multicolumn{2}{@{}l}{\quad \textit{3D (MiniWorld)}} \\
  \addlinespace[1pt]
  CollectHealth          & Collect health packs in a first-person 3D room while avoiding lava. \\
  FourRooms3D            & Navigate through four connected 3D rooms to reach the red target box. \\
  Hallway                & Reach the red box at the end of a long 3D hallway. \\
  Maze                   & Find the red box in a procedurally generated 3D maze. \\
  MazeS2                 & Navigate a small (2$\times$2) 3D maze to find the red box. \\
  MazeS3 / MazeS3Fast    & Navigate a medium (3$\times$3) 3D maze to find the red box. \\
  OneRoom                & Reach the red box in a single open 3D room. \\
  OneRoomS6 / OneRoomS6Fast & Reach the red box in a large (6$\times$6) open 3D room. \\
  PickupObjects          & Find and pick up all boxes matching a target color in a 3D environment. \\
  PutNext                & Pick up a movable box and place it next to the target box in 3D. \\
  RoomObjects            & Navigate around obstacles in a 3D room to reach the red box. \\
  Sidewalk               & Follow a sidewalk path to the end while avoiding obstacles in 3D. \\
  Sign                   & Read a directional sign and navigate to the indicated location in 3D. \\
  TMaze / TMazeLeft / TMazeRight & Navigate a T-shaped 3D maze and choose the correct branch. \\
  ThreeRooms             & Navigate through three connected 3D rooms to reach the target. \\
  WallGap                & Find and navigate through a gap in a 3D wall to reach the goal. \\
  YMaze / YMazeLeft / YMazeRight & Navigate a Y-shaped 3D maze and choose the correct branch. \\
  \midrule
  \multicolumn{2}{@{}l}{\textit{\textbf{Temporal} (13 environments)}} \\
  \addlinespace[2pt]
  Airstriker             & Vertical-scrolling shooter; destroy enemy aircraft with continuous fire and evasion. \\
  AlteredBeast           & Side-scrolling beat-em-up; fight enemies and collect power-ups to transform. \\
  CastleOfIllusion       & Side-scrolling platformer; navigate levels, defeat enemies, and rescue the princess. \\
  CastlevaniaBloodlines  & Side-scrolling action platformer; fight through stages of gothic enemies and bosses. \\
  Columns                & Falling-block puzzle; match three or more same-colored jewels vertically, horizontally, or diagonally. \\
  DynamiteHeaddy         & Side-scrolling platformer; swap heads for different abilities to progress through levels. \\
  GoldenAxe              & Side-scrolling hack-and-slash; fight through enemies using melee attacks and magic. \\
  KidChameleon           & Side-scrolling platformer; collect helmets granting different transformation abilities. \\
  MortalKombatII         & Fighting game; defeat opponents using martial arts moves and special attacks. \\
  SpaceHarrierII         & Third-person rail shooter; dodge obstacles and destroy enemies while auto-scrolling. \\
  StreetsOfRage2          & Side-scrolling beat-em-up; fight through urban stages using combos and special moves. \\
  Strider                & Side-scrolling action game; slash through enemies with a plasma sword across global stages. \\
  ThunderForceIII        & Horizontal-scrolling shooter; destroy enemy waves and bosses with upgradeable weapons. \\
\end{longtable}

\end{document}